\documentclass[]{fairmeta} 
\usepackage{fontawesome5} %
\usepackage{twemojis}     %
\definecolor{dmGreen}{HTML}{00693E}  %
\definecolor{dmForest}{HTML}{12312B} %
\colorlet{metabg}{dmGreen!12}        %
\colorlet{metafg}{dmForest}          %

\usepackage{amsmath,amsfonts,bm}

\def\eqref#1{equation~\ref{#1}}

\def\1{\bm{1}}

\DeclareMathAlphabet{\mathsfit}{\encodingdefault}{\sfdefault}{m}{sl}
\SetMathAlphabet{\mathsfit}{bold}{\encodingdefault}{\sfdefault}{bx}{n}

\usepackage{url}
\usepackage{xcolor}
\usepackage{booktabs}
\usepackage{multirow}
\usepackage{graphicx}
\usepackage{pifont}
\usepackage{colortbl}

\colorlet{headergreen}{dmGreen!12} %
\definecolor{textred}{HTML}{980000}
\definecolor{mygray}{gray}{0.5}
\definecolor{textgreen}{HTML}{2E7D32}
\definecolor{topone}{HTML}{08306B}
\definecolor{toptwo}{HTML}{6BAED6}
\definecolor{cReasoning}{HTML}{17807E}  %
\definecolor{cTimeSeries}{HTML}{8C510A} %
\definecolor{cQuestion}{HTML}{6E6E6E}   %
\definecolor{cAnswer}{HTML}{8C4A96}     %
\definecolor{cGlobalDesc}{HTML}{E2E8F0} %
\definecolor{cLocalDesc}{HTML}{FEFFEE}  %
\definecolor{cInference}{HTML}{E6E6FF}  %
\definecolor{cAnswerBox}{HTML}{CEDBA6}  %
\newcommand{\hlfig}[2]{{\setlength{\fboxsep}{1.5pt}\colorbox{#1}{#2}}}
\definecolor{cDescFill}{HTML}{DFEDF6}   \definecolor{cDescText}{HTML}{205C80}   %
\definecolor{cInferFill}{HTML}{EBE5F6}  \definecolor{cInferText}{HTML}{624697}  %
\definecolor{cAnsFill}{HTML}{E3EFDC}    \definecolor{cAnsText}{HTML}{39663C}    %
\newcommand{\hlfigtext}[3]{{\setlength{\fboxsep}{1.5pt}\colorbox{#1}{\textcolor{#2}{\textbf{#3}}}}}
\newcommand{\negval}[1]{\textcolor{textred}{\scriptsize (#1)}}
\newcommand{\posval}[1]{\textcolor{textgreen}{\scriptsize (#1)}}
\newcommand{\hl}{\cellcolor{headergreen}}

\usepackage{float}        %
\usepackage{wrapfig}      %
\usepackage{hyperref}
\hypersetup{colorlinks=true, linkcolor=red, citecolor=dmGreen, filecolor=magenta, urlcolor=magenta} %
\usepackage{algorithm}    %
\usepackage{listings}     %
\usepackage[T1]{fontenc}  %
\usepackage{upquote}      %
\lstdefinestyle{mocov3}{
  backgroundcolor=\color{white},
  basicstyle=\fontsize{7.5pt}{7.5pt}\ttfamily\selectfont,
  columns=fullflexible,
  breaklines=true,
  captionpos=b,
  commentstyle=\fontsize{7.5pt}{7.5pt}\color[rgb]{0.25,0.5,0.5},
  keywordstyle=\fontsize{7.5pt}{7.5pt}\color[rgb]{0.85,0.18,0.50},
  numbers=none,
}

\newcommand{\sysemoji}{\raisebox{-0.4ex}{\includegraphics[height=2.5ex]{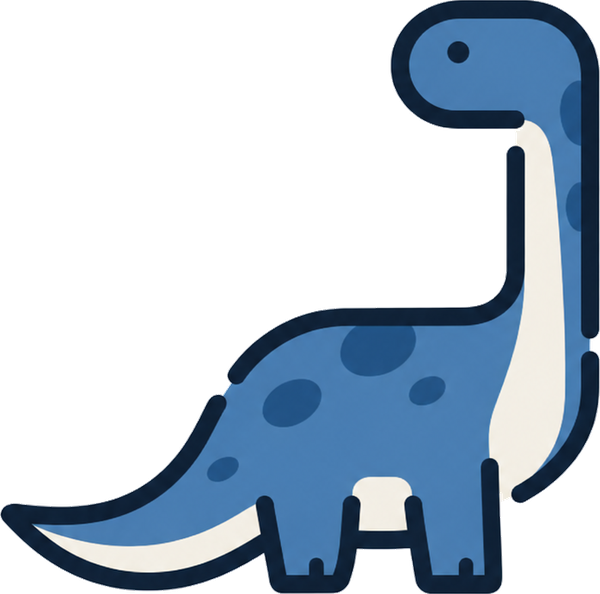}}\hspace{0.3em}}

\title{\sysemoji Lapras: Latent Reasoning for Time Series Language Models}

\author[1]{Yuliang Chen}
\author[1]{Yu Yvonne Wu}
\author[2,3,4]{Patrick Langer}
\author[1]{Arvind Pillai}
\author[1]{Sudarshan Regmi}
\author[2,3]{Martin Maritsch}
\author[5]{Juncheng Liu}
\author[2,3]{Robert Jakob}
\author[2,3,4]{Thomas Kaar}
\author[1]{Tess Z. Griffin}
\author[1]{Lisa Marsch}
\author[1]{Michael V. Heinz}
\author[1]{Nicholas C. Jacobson}
\author[1]{Andrew Campbell}
\affiliation[1]{Dartmouth College}
\affiliation[2]{Aionic Labs}
\affiliation[3]{Agentic Systems Lab, ETH Zurich}
\affiliation[4]{Stanford University}
\affiliation[5]{National University of Singapore}

\abstract{
Time Series Language Models (TSLMs) offer a promising path toward time series understanding by reasoning over temporal signals and producing natural language answers and explanations. A common approach is Chain-of-Thought (CoT), which generates step-by-step rationales describing relevant signal patterns and connecting them to final answers.
Although these models learn from reference CoT traces during post-training, generating faithful descriptions of input time series at inference time remains challenging. The model is required to express high-dimensional and continuous temporal representations in discrete language tokens, which may lead to neglecting task-relevant patterns or describing them inaccurately. Because subsequent reasoning steps build on these descriptions, early errors can propagate through the reasoning chain, ultimately leading to incorrect answers accompanied by plausible explanations that are inconsistent with the input signal.
To address this, we propose \textit{\textbf{\sysname{}} (\textbf{La}tent \textbf{P}ost-trained \textbf{R}easoning \textbf{A}cross \textbf{S}eries)},\footnotemark{} a post-training framework that equips TSLMs with latent reasoning.
A model trained with \sysname{} reasons through a sequence of \emph{continuous thoughts} in the joint time series--language space, producing text only for the final answer. It learns this through teacher-student self-distillation, where a teacher trained on CoT reference traces reasons explicitly through text.
The student is trained to align its hidden states with the teacher's at the answer stage, transferring the teacher's reasoning ability into the student's latent computation.
We evaluate \sysname{} across four TSLM backbones on five diverse time series question answering benchmarks. \sysname{} improves average F1 by up to 10.79\% over explicit CoT while generating $23.9\times$ fewer tokens. 
Importantly, \sysname{}'s generated continuous thoughts can be decoded into readable reasoning traces through standard language decoding, preserving the capability to provide textual explanations.
Together, these results highlight \sysname{} as a promising post-training paradigm for TSLMs that enables efficient and effective reasoning at inference while supporting interpretability.
}

\correspondence{\email{yuliang.chen.gr@dartmouth.edu}}
\metadata[{\faGithub\ Code}]{\url{https://github.com/yuc0805/Lapras}}
\metadata[{\twemoji{smiling face with open hands}\ Hugging Face}]{\url{https://huggingface.co/leochen085/Lapras}}

\newcommand{\yw}[1]{}
\newcommand{\lc}[1]{}
\newcommand{\ap}[1]{}
\newcommand{\sysname}{Lapras}

\begin{document}

\maketitle
\footnotetext{Named after the Pok\'emon Lapras, which understands human speech, though it does not speak, and ferries people across the sea on its back. Likewise, a TSLM post-trained with \sysname{} learns from reference CoT traces written in natural language but does not verbalize its intermediate reasoning at inference: it carries the reasoning from the input time series to the answer through continuous thoughts in the joint time series--language space, and generates text only for the final answer.}

\section{Introduction}
Time Series Language Models (TSLMs) treat time series as a native modality for LLMs, enabling reasoning over sensor signals in natural language across domains ranging from wearable health to industrial monitoring \citep{chatts,itformer}.
To develop reasoning capability, existing works mainly elicit reasoning through Chain-of-Thought (CoT) supervised finetuning (SFT), a common post-training method that further finetunes pretrained models on reference CoT traces~\citep{opentslm}.
After training, the model generates its own step-by-step rationale before producing the answer, providing a textual explanation that users can inspect against the underlying signal.
For example, for the ECG question answering task, a TSLM can generate rationales describing observed waveforms such as an irregular R-R interval and explain how these are linked to a diagnostic answer like candidate arrhythmia~\citep{xu2026multimodalmodelsreasonecg}. 

However, learning from reference CoT traces during post-training does not ensure that models generate traces faithful to the input time series at inference~\citep{turpin2023languagemodelsdontsay}. Generating a CoT requires the model to express high-dimensional, continuous time series representations as discrete language tokens at each step. The resulting description may neglect temporal patterns or characterize them inaccurately.
Because subsequent reasoning steps are predominantly conditioned on this description rather than on the original continuous representation, early omissions or errors can propagate and compound throughout the reasoning chain, ultimately leading to an incorrect answer~\citep{zhang2023languagemodelhallucinationssnowball}.
For instance, for a TSR question over a temperature signal, a generated CoT trace may correctly identify an upward trend and a peak but overlook that the increase is temporary.
Subsequent reasoning may then interpret this temporary increase as a sustained abnormal event and arrive at an incorrect operational decision, such as activating emergency backups when standard capacity is sufficient~(Figure~\ref{fig:motivation}).
We term this limitation the \emph{verbalization bottleneck}: generated textual descriptions may omit or misrepresent task-relevant temporal patterns, and these errors can propagate through subsequent reasoning.
Our analysis in Section~\ref{sec:gap} provides empirical support for this bottleneck, showing that a substantial gap in answer accuracy emerges between predicted and reference CoT traces during signal description and persists through subsequent reasoning.

Recent latent reasoning frameworks offer a potential way to address this bottleneck by reasoning in continuous hidden states rather than generating intermediate text~\citep{coconut,codi}. At each reasoning step, the model's hidden state is passed directly as input to the next step, replacing the text tokens used in CoT.
Although these methods have demonstrated promising results on text-based reasoning tasks, where both inputs and reasoning traces share the same discrete token space, their effectiveness for reasoning over the continuous time series representation space remains underexplored.
Consequently, we ask: \emph{can TSLMs reason more effectively in a continuous latent space than through discrete tokens?}

Motivated by this question, we propose~\textit{\textbf{\sysname{}} (\textbf{La}tent \textbf{P}ost-trained \textbf{R}easoning \textbf{A}cross \textbf{S}eries)}, a post-training framework for latent reasoning in TSLMs.
Specifically, we adopt the self-distillation framework~\citep{codi}, which lets the model learn from the reasoning signal in explicit CoT traces during training without having to verbalize numerical patterns as text at inference.
When training with \sysname{}, the teacher processes the time series, question, and explicit CoT reference traces, while the student reasons over the same time series and question through continuous thoughts.
The student learns to match the teacher's hidden states at the answer position, without requiring each continuous thought to generate a textual reasoning step. At inference, the student performs intermediate reasoning in the joint time series--language space and generates text only for the final answer, mitigating the verbalization bottleneck by avoiding intermediate textual descriptions of the signal. Our experiments compare \sysname{} against existing reasoning approaches across different TSLM backbones and demonstrate that \sysname{} improves average answer accuracy over explicit CoT while substantially reducing token generation.
In summary, our key contributions are:
\begin{itemize}
    \item We identify and empirically characterize the \emph{verbalization bottleneck} in TSLMs, showing that prediction inaccuracy for explicit CoT emerges during signal description and persists through reasoning chains (Section~\ref{sec:gap}).
    \item To address this limitation, we propose \sysname{}, which adapts teacher-student self-distillation to post-train TSLMs for latent reasoning, letting the model learn from reference CoT while performing intermediate reasoning through continuous thoughts~(Section~\ref{sec:method}).
    \item Across four TSLM backbones and five reasoning benchmarks, \sysname{} improves average F1 by up to 10.79\% over CoT while substantially reducing the number of generated tokens. Qualitative analyses show that continuous thoughts of \sysname{} can be decoded into readable reasoning traces, retaining explanation capability (Section~\ref{sec:results}).
\end{itemize}

\section{Related Work}

\textbf{Time Series Language Models.} 
While Time Series Foundation Models (TSFMs) excel at forecasting and fixed-label classification, their rigid outputs preclude open-ended reasoning over sensor signals~\citep{chronos,TimesFM,moment,normwear}. Recent works aim to bridge this gap, progressing from serializing raw data as text~\citep{liu2023largelanguagemodelsfewshot,kim2024healthllmlargelanguagemodels,gruver2024largelanguagemodelszeroshot} to encoding signals into continuous representations that the LLM processes directly~\citep{chatts,opentslm,slip}. These models elicit reasoning via explicit CoT through supervised finetuning~\citep{opentslm,artist,timeomni} or reinforcement learning~\citep{artist,timeomni}. However, verbalizing intermediate rationales into discrete tokens may inflate inference latency~\citep{wu2025groundedchainofthoughtmultimodallarge} and risk hallucination due to excessive reliance on linguistic priors~\citep{zhu2026visualcotinternalizedvisual}. In contrast, \sysname{} conducts multi-step reasoning entirely within the model's continuous latent space, sidestepping intermediate token generation while still decoding final answers into free-form text.

\textbf{Implicit Chain-of-Thought Reasoning.} 
Implicit reasoning methods replace discrete decoded tokens with continuous hidden states which are fed back as input embeddings, freeing models from enforcing textual coherence or committing prematurely to a single reasoning path~\citep{icot,coconut}. Rather than verbalizing steps, iCoT~\citep{icot} progressively removes CoT tokens during training to internalize the reasoning process directly into model weights. COCONUT~\citep{coconut} preserves these intermediate positions by feeding the last-layer hidden state back as the next input embedding, which serves as a continuous thought. CODI~\citep{codi} adopts these continuous thoughts and supervises them through self-distillation against an explicit CoT teacher. While these methods do not yet consistently match explicit CoT in accuracy, eliminating intermediate token generation substantially reduces inference latency, establishing latent reasoning as a compelling alternative paradigm. In parallel, recent vision-language models interleave latent visual tokens with text tokens to manipulate visual evidence without generating explicit images~\citep{mirage,lvr,monet}. However, unlike static image patches that carry independent semantic meaning, time series signals may require jointly interpreting transient local events (e.g., peaks) and global properties (e.g., periodicity and cross-channel dependencies) that are inherently difficult to express in natural language. \sysname{}, instead, reasons in a joint time series--language space, directly coupling temporal and spatial dynamics with semantic anticipation.

\section{Method}

\begin{figure}[t]
    \centering
    \includegraphics[width=\linewidth]{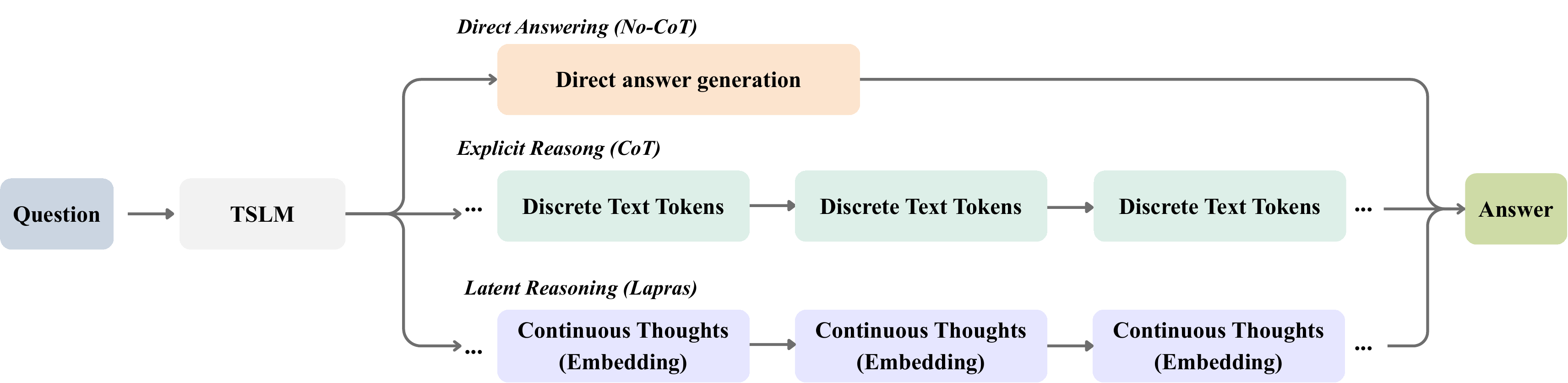}
    \caption{\textbf{Three reasoning paradigms for TSLMs.} \textit{(i) Direct answering (No-CoT)} maps the input directly to a textual answer without generating intermediate reasoning steps. \textit{(ii) Explicit reasoning (CoT)} expresses intermediate reasoning as discrete text tokens and conditions on these steps to produce the answer. \textit{(iii) Latent reasoning (\sysname{})} performs intermediate reasoning through continuous thoughts in latent space and decodes text only for the final answer.}
    \label{fig:post_train}
\end{figure}

\subsection{Diagnosing the Verbalization Bottleneck}
\label{sec:gap}
In time series question answering, a TSLM takes a time series and a question as input and generates a textual answer, either directly or through intermediate reasoning steps (Figure~\ref{fig:post_train}).
Explicit CoT requires a TSLM to describe temporal evidence in language and condition subsequent reasoning on these descriptions. Inaccurate or incomplete descriptions may therefore affect the final answer (i.e., \textit{verbalization bottleneck}). To investigate this issue, we examine how answer accuracy evolves as the model receives successive steps of predicted or reference CoT traces. Specifically, we investigate whether an accuracy gap emerges during signal description and whether subsequent reasoning closes this gap.

\textbf{Setup.}
We adapt the \emph{early answering} protocol from \citet{lanham2023measuring}, which tests how a model's final answer depends on its predicted CoT traces. We apply it to SLIP-1B finetuned with CoT-SFT. Specifically, we progressively truncate both predicted and reference reasoning traces one sentence (hereafter, a \emph{step}) at a time, prompting the model for a final answer after each step to track accuracy. Based on manual inspection, we assign each sentence of the CoT to one of two stages: \emph{describe}, where it summarizes the input signal in words, and \emph{reason}, where it draws conclusions from the preceding description.
These annotations allow us to examine when answer accuracy diverges between the two conditions and how the gap changes through subsequent reasoning.

\begin{wrapfigure}{r}{0.5\linewidth}
    \centering
    \includegraphics[width=\linewidth]{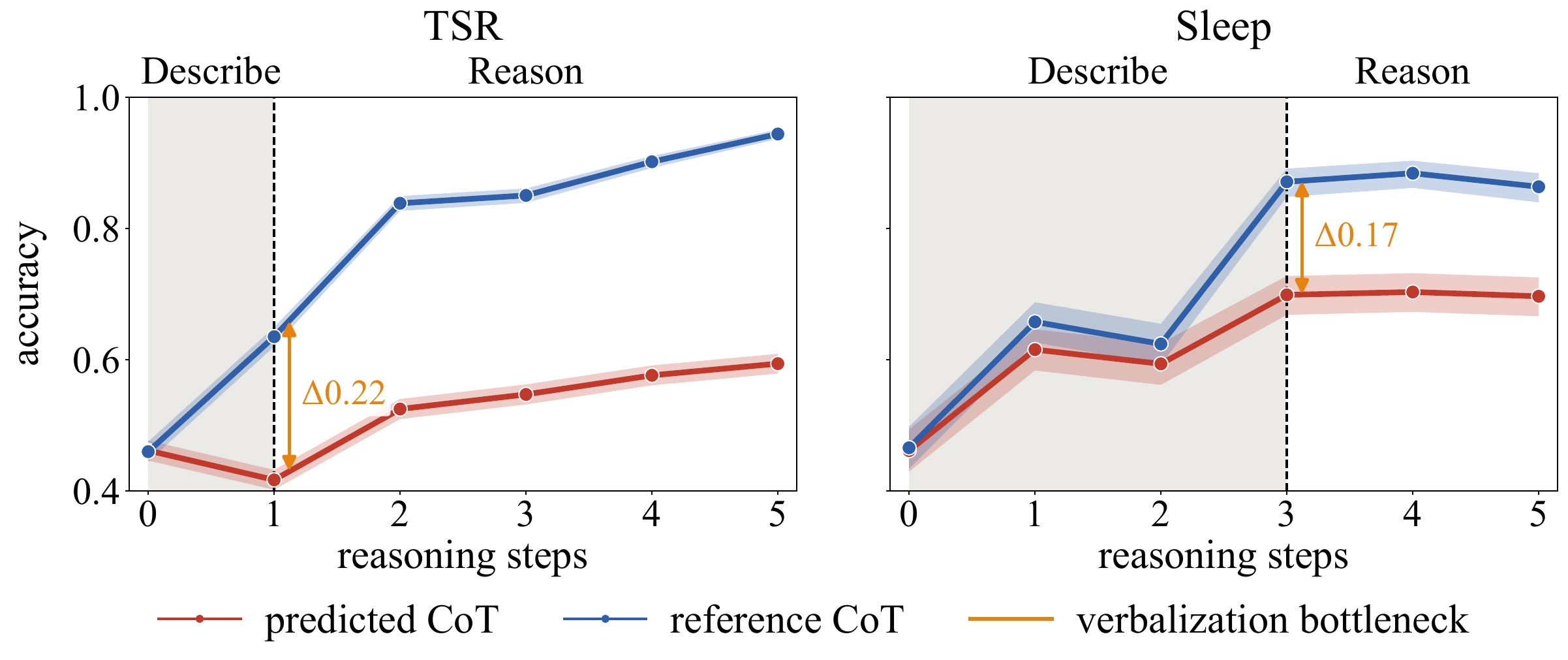}
    \caption{\textbf{Early answering with predicted and reference CoT.}
    A substantial accuracy gap between predicted and reference traces is evident by the end of signal description and persists through subsequent reasoning.
    Mean and standard deviation are computed over the full test set of each benchmark.}
    \label{fig:verbalization_loss}
\end{wrapfigure}
\begin{figure}[h]
    \centering
    \includegraphics[width=1\linewidth]{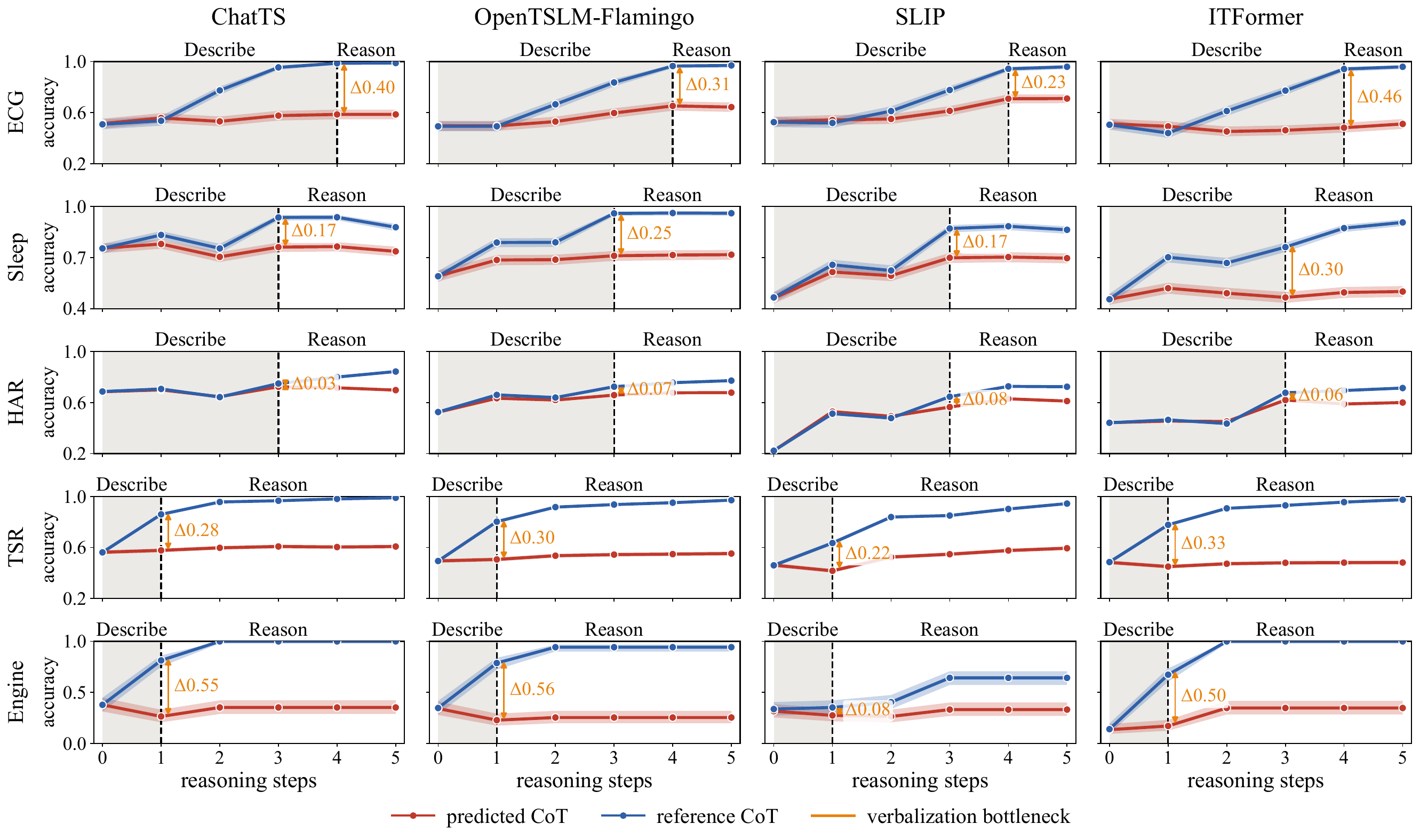}
    \caption{\textbf{Early-answering accuracy under predicted and reference CoT.} Accuracy of a frozen CoT-SFT model answering from only the first $k$ CoT sentences, using either its own predicted CoT or the reference annotation. The verbalization bottleneck $\Delta$ is the accuracy difference between the two at the end of the \emph{describe} stage. Shaded regions mark the \emph{describe} and \emph{reason} stages.}
    \label{fig:app_verbalization_gap}
\end{figure}

\textbf{Results and Implications.}
As shown in Figure~\ref{fig:verbalization_loss}, a substantial performance gap emerges in the \emph{describe} stage, reaching $\Delta 0.22$ on TSR and $\Delta 0.17$ on Sleep. This gap persists through the final \emph{reason} stage, where the predicted trace's accuracy plateaus. More results are provided in Figure~\ref{fig:app_verbalization_gap}.
These findings provide empirical support for the \emph{verbalization bottleneck} where translating continuous time series into text during the describe stage discards task-relevant evidence, introducing errors that propagate through the reasoning chain. Moreover, this analysis offers a possible explanation for why CoT-SFT struggles to outperform No-CoT baselines (Table~\ref{tab:main_results}). Ultimately, this motivates learning from reference rationales while reasoning in continuous space, eliminating the need to generate text descriptions at inference.

\subsection{\sysname{}}
\label{sec:method}

\begin{figure}[t]
    \centering
    \includegraphics[width=1\linewidth]{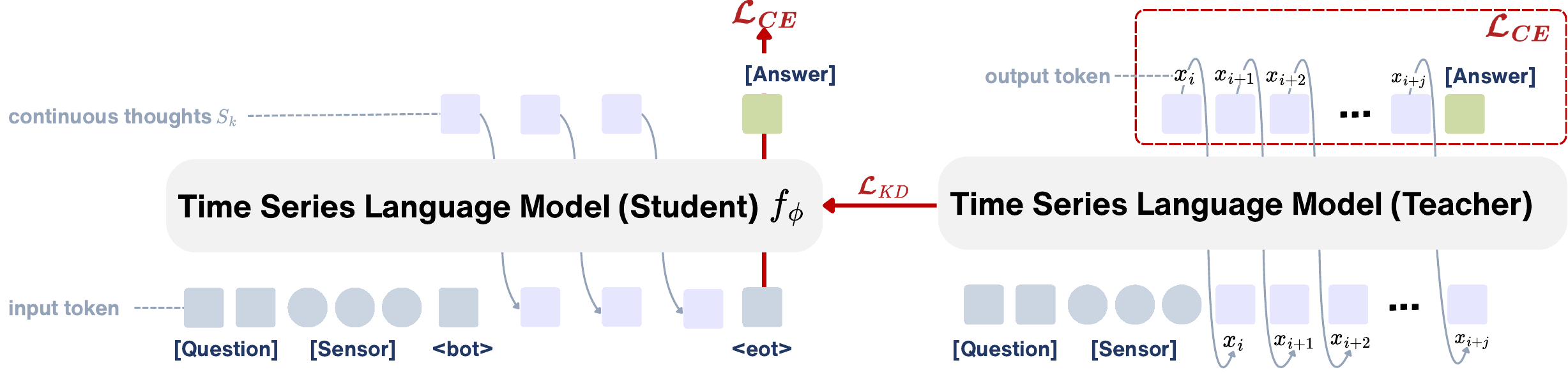}
    \caption{\textbf{Training framework of \sysname{}.} The student (left) produces $K$ continuous thoughts between $\langle\mathrm{bot}\rangle$ and $\langle\mathrm{eot}\rangle$, feeding each hidden state back as the next input embedding without emitting any token, and predicts the answer under $\mathcal{L}^{\mathrm{student}}_{\mathrm{ans}}$. The teacher (right) writes a full CoT~(e.g., $[x_i, x_{i+1}, \dots, x_{i+j}]$) and predicts the answer under $\mathcal{L}^{\mathrm{teacher}}_{\mathrm{CE}}$. A distillation loss $\mathcal{L}_{\mathrm{KD}}$ aligns their hidden states at the anchor position (the first answer token). The teacher is discarded at inference.}
    \label{fig:method}
\end{figure}

\textbf{Overview.} To address the verbalization bottleneck, we introduce \sysname{} (Latent Post-trained Reasoning Across Series), which reasons over the question and time series through continuous thoughts in latent space, decoding text only for the final answer. As shown in Figure~\ref{fig:method}, \sysname{} adopts a knowledge distillation framework~\citep{dino,codi}. Here, a teacher network is trained on CoT reasoning traces, while a student network learns to match the teacher's hidden states through latent reasoning. Both the student and teacher models share the same TSLM architecture and parameters. Formally, given a multivariate sensor series $X_s$ and a text prompt $X_q$ (comprising context and a question), a TSLM generates a textual answer $X_a$. We denote the time series representation by $H_s$ and the prompt embeddings by $H_q = \mathrm{embed}(X_q)$. The language model $f_\phi$ conditions on both inputs, integrating $H_s$ through the backbone's existing architecture, and returns the last hidden state at the final position. The answer is decoded autoregressively as $X_a \sim p_\phi(\cdot \mid H_s, H_q)$.

\textbf{Teacher-Student Self-distillation.}
The teacher and student share the same TSLM architecture but perform intermediate reasoning differently. Given the time series and question, the teacher is trained with standard CoT-SFT: it processes the reference CoT trace and is trained with token-level cross-entropy over the traces and answers, $\mathcal{L}^{\mathrm{teacher}}_{\mathrm{CE}}$. The student instead generates continuous thoughts and is trained with cross-entropy over only the answer tokens, $\mathcal{L}^{\mathrm{student}}_{\mathrm{ans}}$. To transfer information from the reference rationale to the student's latent reasoning, we align the hidden states used to predict the first answer token.

\sysname{} introduces two learned markers, $\langle\mathrm{bot}\rangle$
(\textit{begin-of-thought}) and $\langle\mathrm{eot}\rangle$ (\textit{end-of-thought}),
which delimit the latent reasoning process. Given a prompt $X_q$ and a series $X_s$,
the student embeds both 
and appends
$\langle\mathrm{bot}\rangle$, which puts it in \emph{latent mode},
\begin{equation}
\begin{alignedat}{2}
  &S_0 = [\,H_q,\, H_s,\, \langle\mathrm{bot}\rangle\,], && \\
  &z_k = f_\phi(S_{k-1}),                                &\qquad& S_k = [\,S_{k-1};\, \pi(z_k)\,], \\
  &X_a \sim p_\phi(\cdot \mid [\,S_K;\, \langle\mathrm{eot}\rangle\,]), &&
\end{alignedat}
  \label{eq:latent_loop}
\end{equation}
where $\pi$ is a learned projection from the last hidden state back into the input embedding space and $k = 1, \dots, K$.
In latent mode the student never samples a token. Instead, after each forward pass, it extracts the last hidden state $z_k$ at the final position and projects it back into the input embedding space via a small learned MLP $\pi$. The projected vector is then appended to the input sequence as if it were a new token embedding, and the model runs another forward pass. We refer to $z_1,\dots,z_K$ as \emph{continuous thoughts}~\citep{coconut,codi}: intermediate reasoning representations that encode the model's reasoning process without being decoded into language. After generating $K=6$ continuous thoughts, the student returns to \emph{language mode} at $\langle\mathrm{eot}\rangle$ to decode the final answer $X_a$ in language tokens. $X_a$ is supervised with $\mathcal{L}^{\mathrm{student}}_{\mathrm{ans}}$, the cross-entropy over the answer tokens.

\textbf{Training \& Inference.}
Student and teacher are trained jointly, and the distillation couples them at the
\emph{anchor}, the first answer token of each, where we align their hidden states over
all layers,
\begin{equation}
  \mathcal{L}_{\mathrm{KD}}
  = \frac{1}{M+1}\sum_{\ell=0}^{M}
    \Big|\,\mathrm{sg}\big[h^{\ell}_{\mathrm{teacher}}\big]
          - h^{\ell}_{\mathrm{student}}\,\Big|,
  \label{eq:distill}
\end{equation}
where $M$ is the number of transformer layers, $h^{\ell}$ is the hidden activation of the $\ell$-th layer of the TSLM, and $\mathrm{sg}$ denotes the stop-gradient operation. 
Combining this term with the two cross entropies gives the full objective
$\mathcal{L} = \mathcal{L}^{\mathrm{student}}_{\mathrm{ans}} + \lambda\,\mathcal{L}_{\mathrm{KD}}
+ \beta\,\mathcal{L}^{\mathrm{teacher}}_{\mathrm{CE}}$, which supervises the student both on the answer it produces and on the state it passes through on the way to that answer.
The distillation weight $\lambda$ sets how strongly the student is pulled toward the teacher's states relative to its own answer loss, and we set it to $\lambda=10$ by default and adjust it per task as reported in Table~\ref{tab:hparams}. The teacher weight $\beta$ scales the teacher's own cross entropy, and we keep it at $\beta=1$ throughout.

During training, we apply a standard input regularizer that masks a fraction ($\rho{=}0.3$) of the sensor patches~\citep{TimesFM}. Our ablation study shows that this improves performance by 1.80\% on average (see Table~\ref{tab:ablation}). At inference, we discard the teacher branch and run only the student network (Figure~\ref{fig:method}, left) through the following steps. First, the student ingests the prompt $X_q$ and sensor series $X_s$. Second, it performs $K$ latent reasoning steps as described in Equation~\ref{eq:latent_loop}, emitting no intermediate tokens. Finally, after appending $\langle\mathrm{eot}\rangle$, the student transitions back to language generation and decodes the final answer $X_a$ conditioned on $[\,S_K;\, \langle\mathrm{eot}\rangle\,]$.

\vspace{-1mm}
\section{Experiment}
\vspace{-2mm}
\label{sec:experiment}
\textbf{Datasets.}
\label{sec:data}
We evaluate \sysname{} on five time series question answering tasks: ECG~\citep{ECG-Reasoning-Benchmark} (cardiological diagnosis), Engine~\citep{itformer} (aero-engine fault diagnosis), HAR~\citep{opentslm} (human activity recognition), Sleep~\citep{opentslm} (sleep-stage classification), and TSR~\citep{merrill-etal-2024-language} (counterfactual prediction).

To support explicit CoT post-training, we construct reference reasoning traces for each task. 
For ECG and Engine, we construct traces using predefined templates populated with their original annotated step labels and short answers.
For HAR, Sleep, and TSR, we generate reference traces using Qwen3.5-27B, following \citet{artist}. Full details and per-dataset statistics are provided in Appendix~\ref{sec:app_dataset} and Table~\ref{tab:datasets}.

\textbf{Baselines.}
We compare \sysname{} against three classes of reasoning approaches. \textbf{(i) No-CoT} produces direct answers without generating reasoning traces. \textbf{(ii) Explicit CoT} generates step-by-step textual rationales before producing final answers, trained on reference reasoning traces. \textbf{(iii) Latent reasoning} includes two baselines: \textbf{iCoT} progressively removes CoT tokens during training so that the model learns to reason without explicit verbalization, and \textbf{COCONUT} progressively replaces textual steps with continuous hidden states through curriculum learning.

\textbf{Backbones and Metrics.} 
We evaluate all reasoning approaches across four TSLM backbones (ChatTS-7B, OpenTSLM-1B, SLIP-1B, and ITFormer-0.5B) on five reasoning tasks, where OpenTSLM-1B refers to its OpenTSLM-Flamingo variant (see Appendix~\ref{sec:app_arch} for the OpenTSLM-SoftPrompt variant). Backbone details are provided in Appendix~\ref{sec:app_backbones}. For evaluation, we parse the predicted class label from each model's generated output and compare it with the ground-truth label. We report accuracy and macro-F1 for each task.

\section{Results and Discussion} \label{sec:results}
\subsection{Main Results} 
\begin{table}[t]
\centering
\caption{\textbf{Main Result.} Accuracy and Macro-F1 (\%) of five finetuning methods on four
TSLMs across five reasoning tasks. Our method is shaded in \colorbox{headergreen}{green}.
\textbf{Bold} and \underline{underline} denote the best and second-best per column
within each model. $\Delta$ reports the change in average relative to No-CoT.
}
\label{tab:main_results}
\resizebox{\linewidth}{!}{%
\begin{tabular}{ll cc cc cc cc cc rl rl}
\toprule
 & & \multicolumn{2}{c}{ECG} & \multicolumn{2}{c}{Sleep} & \multicolumn{2}{c}{HAR} & \multicolumn{2}{c}{TSR} & \multicolumn{2}{c}{Engine} & \multicolumn{4}{c}{Avg.} \\
\cmidrule(lr){3-4} \cmidrule(lr){5-6} \cmidrule(lr){7-8} \cmidrule(lr){9-10} \cmidrule(lr){11-12} \cmidrule(lr){13-16}
Model & Finetune Method & Acc & F1 & Acc & F1 & Acc & F1 & Acc & F1 & Acc & F1 & Acc & $\Delta$ & F1 & $\Delta$ \\
\midrule
\multirow{6}{*}{ChatTS-7B}
 & No-CoT & 81.65 & 81.63 & \underline{87.65} & 78.05 & 69.42 & 63.41 & 66.54 & 66.50 & 27.98 & 27.92 & 66.65 & & 63.50 & \\
 & CoT & 59.25 & 58.56 & 79.63 & 65.48 & 72.79 & 67.27 & 61.90 & 61.87 & 35.23 & \textbf{35.49} & 61.76 & \negval{-4.89} & 57.73 & \negval{-5.77} \\
\cmidrule{2-16}
 & \multicolumn{15}{l}{\color{gray!60}\textbf{\textit{Latent Reasoning:}}} \\
 & iCoT &76.83 & 76.76 & 87.00 & \underline{78.54} & \underline{73.74} & \underline{69.55} & \underline{67.20} & \underline{67.16} & 33.68 & 32.81 & 67.69 & \posval{+1.04} & 64.96 & \posval{+1.46} \\
 & COCONUT & \underline{84.45} & \underline{84.41} & 86.67 & 77.31 & 73.52 & 68.36 & 65.58 & 65.56 & \textbf{36.27} & \underline{35.24} & \underline{69.30} & \posval{+2.65} & \underline{66.18} & \posval{+2.67} \\
 & \hl \sysname{} (Ours) & \hl\textbf{85.23} & \hl\textbf{85.17} & \hl\textbf{88.19} & \hl\textbf{79.49} & \hl\textbf{77.43} & \hl\textbf{73.53} & \hl\textbf{69.32} & \hl\textbf{69.32} & \hl\underline{35.75} & \hl 35.10 & \hl\textbf{71.18} & \hl\posval{+4.54} & \hl\textbf{68.52} & \hl\posval{+5.02} \\
\midrule
\multirow{6}{*}{OpenTSLM-1B}
 & No-CoT & 71.07 & 70.90 & \underline{77.57} & \underline{65.25} & \underline{71.31} & \underline{66.22} & 33.29 & 33.32 & \underline{28.50} & 11.09 & 56.35 & & 49.36 & \\
 & CoT & 65.47 & 65.33 & 72.81 & 58.02 & 67.61 & 61.71 & \underline{52.69} & \underline{53.13} & 27.98 & \underline{26.48} & 57.31 & \posval{+0.96} & 52.93 & \posval{+3.58} \\
\cmidrule{2-16}
 & \multicolumn{15}{l}{\color{gray!60}\textbf{\textit{Latent Reasoning:}}} \\
 & iCoT &\underline{72.94} & \underline{72.78} & \underline{77.57} & 65.22 & 69.98 & 65.40 & 51.10 & 51.08 & 26.94 & 24.20 & \underline{59.71} & \posval{+3.36} & \underline{55.74} & \posval{+6.38} \\
 & COCONUT & 70.45 & 70.38 & 76.27 & 62.80 & 70.82 & 65.89 & 47.92 & 41.91 & 24.87 & 16.84 & 58.07 & \posval{+1.72} & 51.56 & \posval{+2.21} \\
 & \hl \sysname{} (Ours) & \hl\textbf{73.56} & \hl\textbf{73.49} & \hl\textbf{80.72} & \hl\textbf{68.91} & \hl\textbf{72.40} & \hl\textbf{68.20} & \hl\textbf{58.40} & \hl\textbf{58.44} & \hl\textbf{32.12} & \hl\textbf{27.51} & \hl\textbf{63.44} & \hl\posval{+7.09} & \hl\textbf{59.31} & \hl\posval{+9.95} \\
\midrule
\multirow{6}{*}{SLIP-1B}
 & No-CoT & 71.38 & 71.35 & \underline{77.57} & \underline{64.36} & \underline{73.74} & \underline{69.20} & 50.90 & 50.78 & \underline{37.82} & \textbf{34.02} & 62.28 & & \underline{57.94} & \\
 & CoT & 72.01 & 71.87 & 70.53 & 55.88 & 70.49 & 66.71 & \underline{62.16} & \underline{62.17} & 34.20 & \underline{31.54} & 61.88 & \negval{-0.40} & 57.63 & \negval{-0.31} \\
\cmidrule{2-16}
 & \multicolumn{15}{l}{\color{gray!60}\textbf{\textit{Latent Reasoning:}}} \\
 & iCoT &\underline{81.03} & \underline{80.96} & 77.36 & 62.76 & 72.70 & 68.71 & 61.58 & 61.58 & 22.28 & 13.26 & \underline{62.99} & \posval{+0.71} & 57.45 & \negval{-0.49} \\
 & COCONUT & 79.32 & 79.24 & 77.25 & 62.79 & 72.54 & 68.76 & 50.27 & 50.27 & 28.50 & 11.09 & 61.58 & \negval{-0.71} & 54.43 & \negval{-3.51} \\
 & \hl \sysname{} (Ours) & \hl\textbf{82.74} & \hl\textbf{82.64} & \hl\textbf{78.12} & \hl\textbf{66.31} & \hl\textbf{74.84} & \hl\textbf{70.75} & \hl\textbf{67.32} & \hl\textbf{67.30} & \hl\textbf{38.86} & \hl 26.49 & \hl\textbf{68.38} & \hl\posval{+6.09} & \hl\textbf{62.70} & \hl\posval{+4.76} \\
\midrule
\multirow{6}{*}{ITFormer-0.5B}
 & No-CoT & \underline{52.57} & 48.43 & \underline{62.41} & \underline{35.35} & 67.28 & \textbf{59.23} & 48.90 & 47.22 & \underline{35.23} & \underline{32.19} & \underline{53.28} & & \textbf{44.48} & \\
 & CoT & 50.70 & \textbf{50.27} & 53.09 & 29.81 & 64.47 & 56.06 & 48.90 & \underline{48.89} & \textbf{35.75} & \textbf{35.73} & 50.58 & \negval{-2.70} & \underline{44.15} & \negval{-0.33} \\
\cmidrule{2-16}
 & \multicolumn{15}{l}{\color{gray!60}\textbf{\textit{Latent Reasoning:}}} \\
 & iCoT &50.86 & 37.75 & 45.40 & 12.49 & 66.40 & \underline{59.09} & \textbf{55.98} & \textbf{55.98} & 26.94 & 26.17 & 49.12 & \negval{-4.16} & 38.30 & \negval{-6.19} \\
 & COCONUT & 51.48 & 34.55 & 45.50 & 12.51 & \textbf{67.54} & 55.92 & 47.39 & 42.41 & 31.61 & 31.03 & 48.70 & \negval{-4.57} & 35.28 & \negval{-9.20} \\
 & \hl \sysname{} (Ours) & \hl\textbf{53.81} & \hl\underline{49.29} & \hl\textbf{63.06} & \hl\textbf{36.95} & \hl\underline{67.32} & \hl\underline{59.09} & \hl\underline{49.51} & \hl 48.34 & \hl\textbf{35.75} & \hl 26.46 & \hl\textbf{53.89} & \hl\posval{+0.61} & \hl 44.03 & \hl\negval{-0.46} \\
\bottomrule
\end{tabular}%
}
\end{table}

To evaluate the effectiveness of \sysname{}, we compare it with existing reasoning approaches on five sensor question answering datasets, where each task is formulated as multiple-choice classification over sensor time series. Table~\ref{tab:main_results} reports accuracy and F1 for the five settings on four TSLM backbones ranging from 0.5B to 7B parameters. 
Generally, latent reasoning approaches improve performance compared with explicit CoT. In particular, \sysname{} achieves the highest average F1 on three of the four backbones. Compared with explicit CoT, \sysname{} improves F1 by 5.53\% on average across the 20 backbone-dataset configurations, with the largest gain in average F1 of 10.79\% on ChatTS-7B. We also note that \sysname{} achieves higher average F1 than both iCoT and COCONUT on every backbone. Its average F1 advantage
over COCONUT is 6.78\% across the four backbones, demonstrating the effectiveness of self-distillation compared to unconstrained continuous thought.

In addition, CoT underperforms No-CoT in average F1 on three of the four backbones, with an average drop of 2.14\% on these backbones.
This confirms the verbalization bottleneck identified in Section~\ref{sec:gap}: forcing the model to describe fine-grained temporal patterns in text degrades the final answer F1.
We further note that for ITFormer-0.5B, all reasoning methods including \sysname{} underperform No-CoT in average F1, suggesting that a 0.5B-parameter model cannot effectively learn from CoT supervision, consistent with findings by \citet{li2025smallmodelsstrugglelearn}.
Taken together, \sysname{} consistently outperforms explicit CoT and existing latent reasoning methods across backbones and tasks, establishing it as a practical post-training reasoning paradigm for TSLMs.

\subsection{Qualitative Analysis}

\begin{figure}[t]
    \centering
    \includegraphics[width=\linewidth]{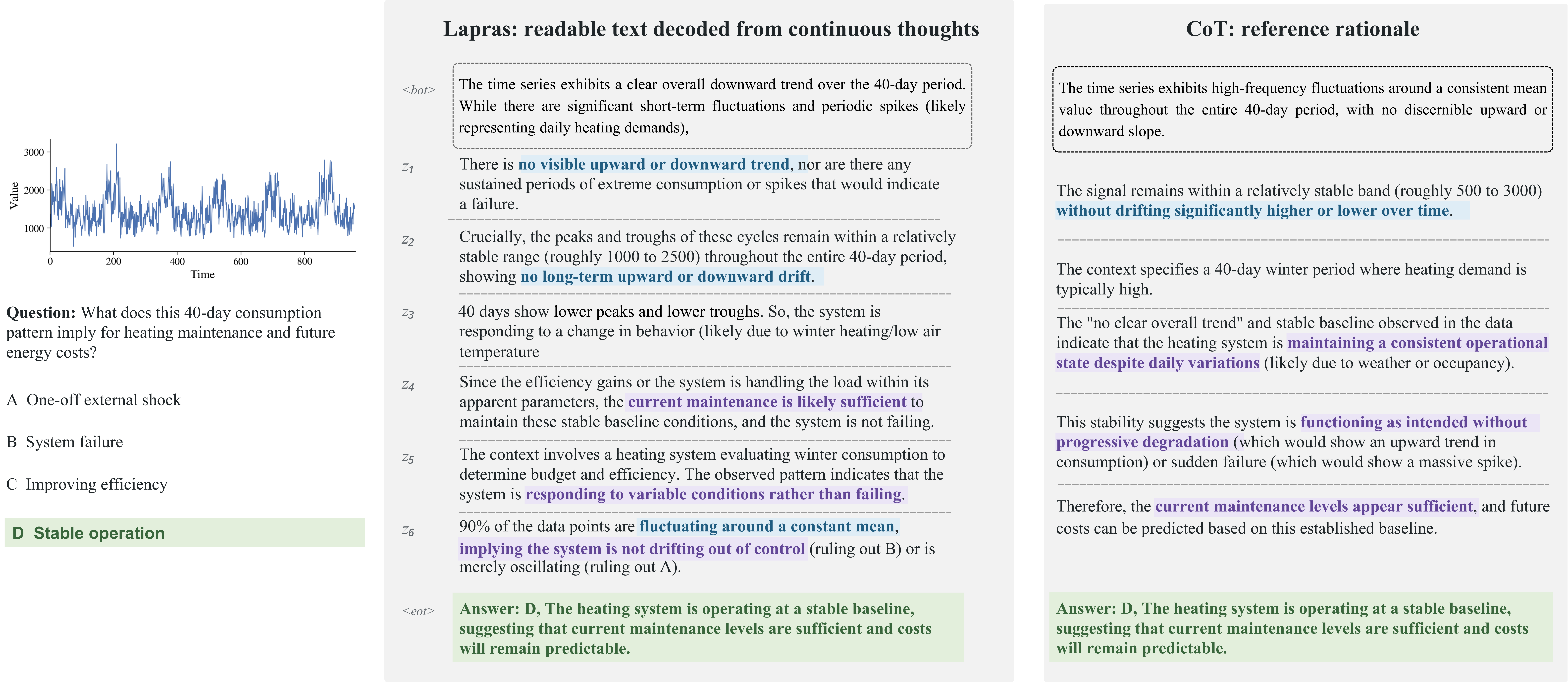}
    \caption{\textbf{Case study of decoded continuous thoughts of \sysname{}.} \sysname{}'s continuous thoughts yield readable, task-relevant text whose content can be compared with the reference rationale.
    The decoded steps progress from a \hlfigtext{cDescFill}{cDescText}{signal description} of the series, through \hlfigtext{cInferFill}{cInferText}{inferences} built on it, to the \hlfigtext{cAnsFill}{cAnsText}{final answer}. Highlight colors match those in the figure.
    }
    \label{fig:interp_718}
\end{figure}

\textbf{\sysname{}'s continuous thoughts can be decoded into readable reasoning traces.}
An advantage of explicit CoT reasoning is that it expresses intermediate steps in natural language, allowing users to inspect the model's reasoning trace. For example, predictions from models deployed in high-stakes settings such as clinical monitoring need human verification before being acted upon. In latent reasoning models, however, these steps are represented as continuous hidden states and are not directly readable. We therefore investigate whether \sysname{} retains the ability to provide human-readable reasoning traces despite performing its intermediate computation in latent space.
Benefiting from the knowledge distillation of \sysname{}, which directly supervises the continuous thoughts, we show that the continuous thoughts can be decoded into human-readable reasoning traces through standard LM decoding with the shared LM head.
Specifically, we greedily decode each position $\langle\mathrm{bot}\rangle, z_1,\dots,z_6, \langle\mathrm{eot}\rangle$ with the LM head.
Figure~\ref{fig:interp_718} shows that the decoded steps contain global series descriptions, local segment descriptions, and inferences that lead to the answer, consistent with the reference reasoning trace. 
This suggests that readable traces can be retrieved post hoc from continuous thoughts, providing a way to audit the model's latent reasoning. Appendix~\ref{sec:app_vis} reports additional cases.

\begin{figure}[t]
    \centering
    \includegraphics[width=0.8\linewidth]{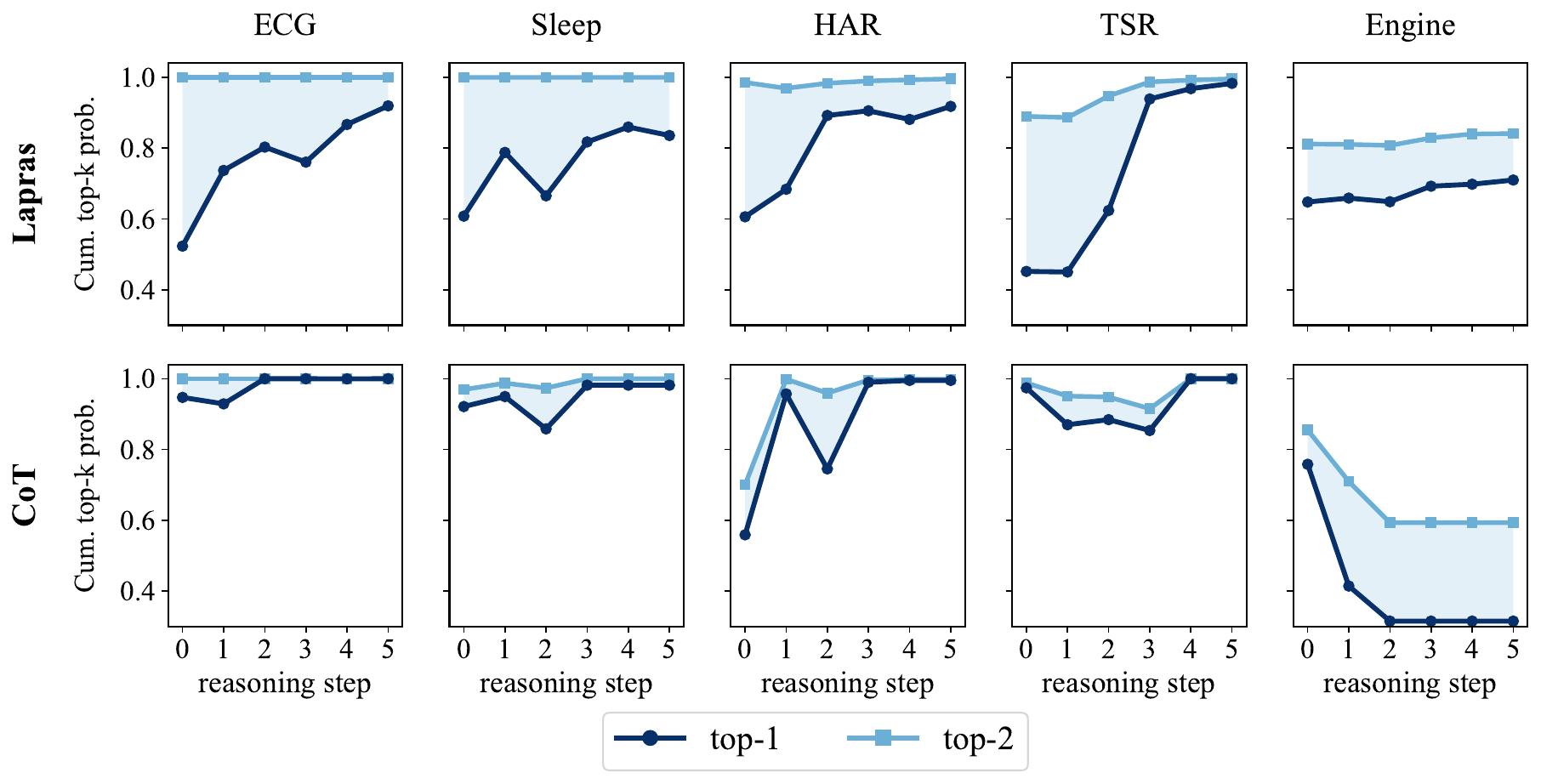}
    \caption{\textbf{Answer probability across reasoning steps} on the SLIP-1B backbone for \sysname{} (top) and CoT (bottom), shown for one representative example per task. Curves show the cumulative probability of \textcolor{topone}{\textbf{Answer in top-1}} and \textcolor{toptwo}{\textbf{Answer in top-2}}. \sysname{} keeps a second candidate alive well into the chain and commits near the end, whereas CoT tends to commit from the first step.}
    \label{fig:answer-topk}
\end{figure}

\textbf{\sysname{} mitigates the verbalization bottleneck by delaying answer commitment.} 
We hypothesize that the verbalization bottleneck forces premature answer commitment at the first reasoning step, indicated by immediately assigning high probability to a single answer that remains unchanged with additional reasoning steps. Hence, mitigating this bottleneck should delay commitment, keeping probabilities balanced across multiple plausible answers early on before converging on the final answer. To evaluate this hypothesis, we track top-1 (most likely) and top-2 (two most likely) answer confidence across reasoning steps for \sysname{} and CoT, following~\citet{wang2025chainofprobeexaminingnecessityaccuracy}. As shown in Figure~\ref{fig:answer-topk}, \sysname{}'s top-1 probability stays well below the combined top-2 probability mass, indicating that the model considers multiple answers before gradually shifting its probability toward the correct choice. In contrast, CoT commits immediately at the first reasoning step, assigning near-total probability mass to a single answer. Because subsequent steps cannot recover from early descriptive errors, CoT depends entirely on the accuracy of its initial verbalization. This supports our preliminary findings~(Section~\ref{sec:gap}) and the logit-lens analysis in Appendix~\ref{sec:logit-lens-entropy}.

\begin{figure}[t]
\centering
\includegraphics[width=0.95\linewidth]{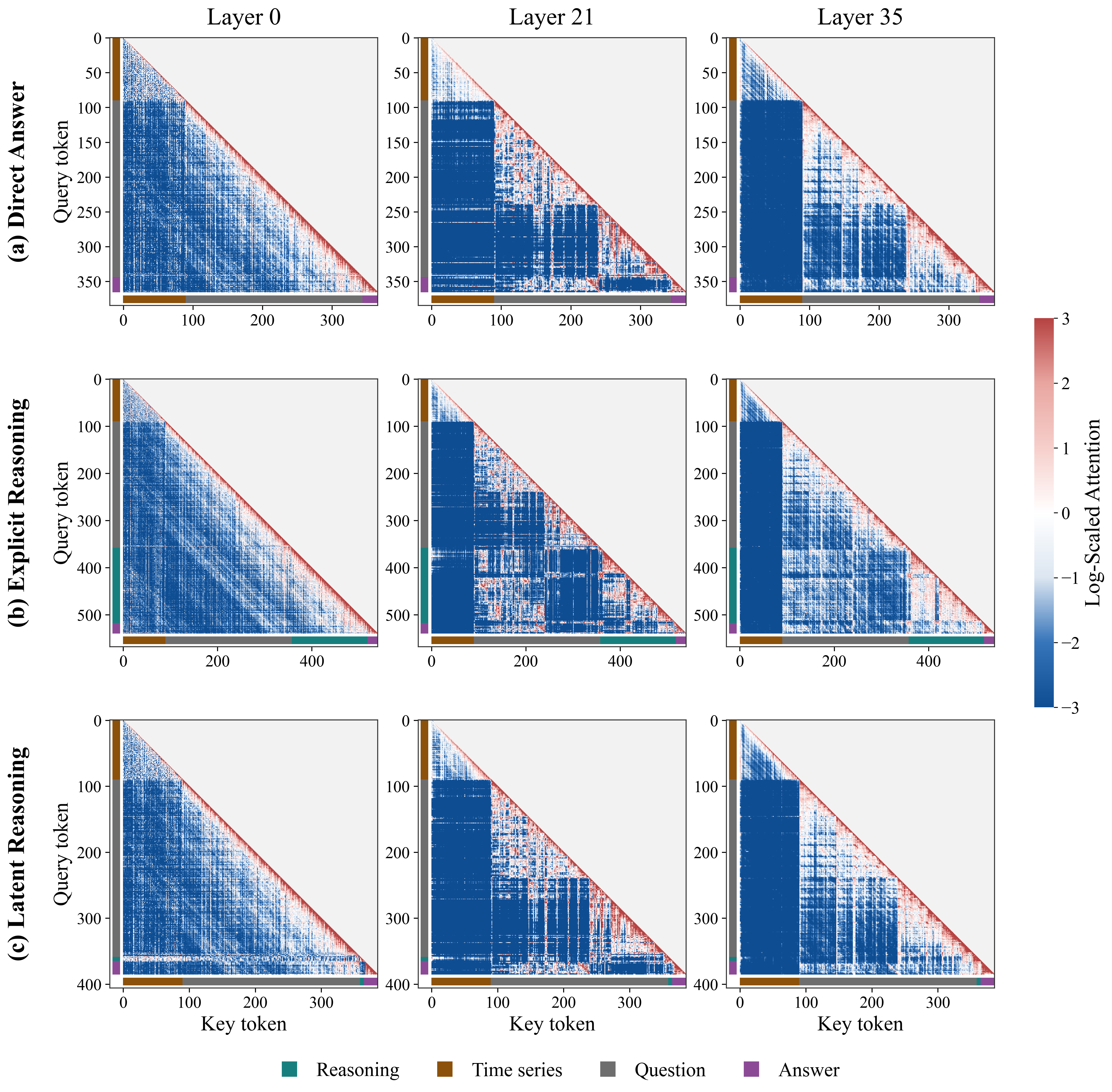}
\caption{\textbf{Attention maps} of ChatTS-7B on TSR at layers $0$, $21$, and
$35$ for \textbf{(a)} No-CoT, \textbf{(b)} explicit chain-of-thought, and
\textbf{(c)} latent reasoning. Color gives attention relative to the uniform level, red
above and blue below, and the bars on the two axes mark the
\textcolor{cTimeSeries}{\textbf{time series}}, \textcolor{cQuestion}{\textbf{question}},
\textcolor{cReasoning}{\textbf{reasoning}}, and \textcolor{cAnswer}{\textbf{answer}} spans.}
\label{fig:attn_map}
\end{figure}

\textbf{\sysname{} tends to allocate more attention to time series representations during reasoning.} To investigate whether avoiding intermediate verbalization encourages greater attention to the continuous temporal representations, we compare the attention patterns of \sysname{} and explicit CoT on a representative TSR example using ChatTS-7B (Appendix~\ref{sec:app_attn}). Figure~\ref{fig:attn_map} and Table~\ref{tab:attn_ts} show that, in this example, \sysname{} assigns a larger share of attention to the time series at all three inspected layers. The difference is largest at the first layer, where \sysname{} allocates 17.35\% of its attention to time series tokens, compared with 3.39\% for CoT. These observations are consistent with \sysname{} drawing more directly on the original temporal representations, while CoT may neglect temporal patterns during verbalization.

\subsection{Quantitative Analysis}

\begin{wrapfigure}{r}{0.55\linewidth}
    \vspace{-4mm}
    \vspace{-\intextsep}
    \centering
    \includegraphics[width=\linewidth]{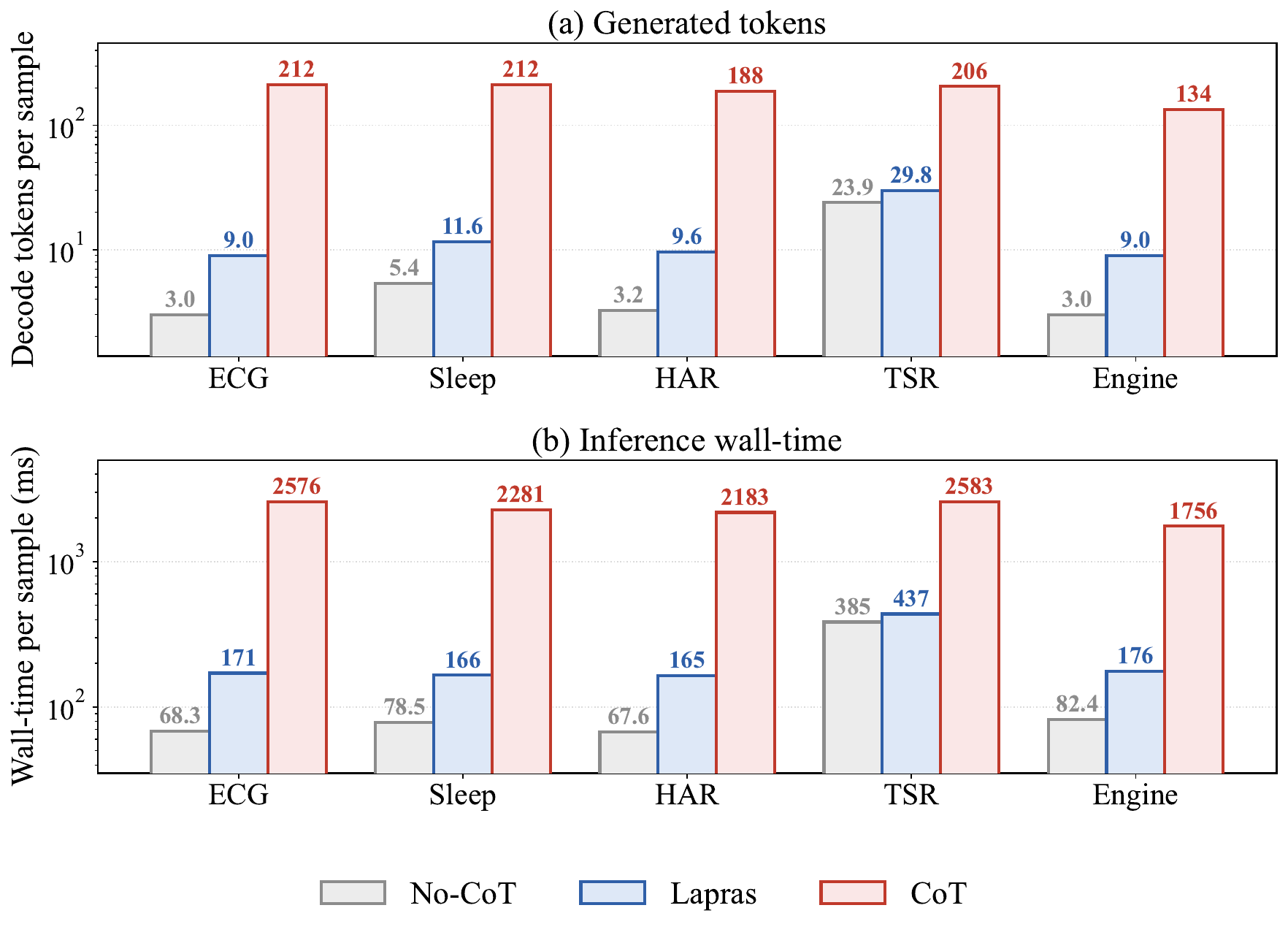}
    \caption{\textbf{Inference efficiency of the three reasoning methods.} Generated tokens and wall-clock time per sample from SLIP-1B with one NVIDIA A6000. 
    }
    \vspace{-4mm}
    \label{fig:efficiency}
\end{wrapfigure}
\textbf{Inference efficiency.}
In CoT, the model autoregressively decodes each reasoning step as text tokens, so inference cost scales linearly with the number of generated tokens. A typical CoT trace can span hundreds of tokens, whereas \sysname{} compresses the entire reasoning process into only $K{=}6$ continuous thoughts, generating text tokens only for the final answer. This results in $6.9$ to $23.9\times$ fewer generated tokens and $5.9$ to $15.1\times$ faster inference than CoT (Figure~\ref{fig:efficiency}), while adding less than $8$\,ms overhead compared to No-CoT. This makes latent reasoning practical for latency-sensitive deployment settings such as real-time health monitoring, where predictions must arrive within strict time constraints.

\textbf{Number of continuous thoughts.}
\sysname{} emits a fixed number of continuous thoughts regardless of the task. Understanding which tasks benefit from these thoughts helps identify where latent reasoning provides value. We test this by taking the model finetuned with $K{=}6$ and varying the number of generated continuous thoughts at inference.
As Table~\ref{tab:num_latent} shows, using 6 continuous thoughts raises average accuracy from $64.82\%$ to $68.38\%$ compared to using a single continuous thought. Increasing to $K{=}8$ does not improve further, with average accuracy slightly declining to $68.07\%$, indicating that additional thoughts beyond the training configuration yield no benefit. The largest gains are on TSR ($+12.46\%$) and Engine ($+4.66\%$). ECG, Sleep, and HAR change by at most $1.09\%$. We attribute this to these tasks depending on local pattern recognition rather than on multi-step reasoning.
This suggests that the benefit of continuous thoughts scales with the amount of multi-step inference a task requires after the time series has been encoded.
\begin{table}[t]
\centering
\caption{\textbf{Number of continuous thoughts.} Answer accuracy (\%) with SLIP-1B, varying only the number of continuous thoughts $K$. Default $K{=}6$ in \colorbox{headergreen}{green}. $\Delta$ reports the change in average relative to $K{=}1$.
}
\label{tab:num_latent}
\resizebox{0.6\linewidth}{!}{%
\begin{tabular}{l ccccc rl}
\toprule
& ECG & Sleep & HAR & TSR & Engine & Avg. & $\Delta$ \\
\midrule
$K{=}1$ & 81.65 & 78.22 & 75.18 & 54.86 & 34.20 & 64.82 & \\
$K{=}3$ & 81.65 & 77.36 & 75.05 & 64.44 & 33.16 & 66.33 & \posval{+1.51} \\
\rowcolor{headergreen}
$K{=}6$ & 82.74 & 78.12 & 74.84 & 67.32 & 38.86 & 68.38 & \posval{+3.56} \\
$K{=}8$ & 82.43 & 77.36 & 74.85 & 67.39 & 38.34 & 68.07 & \posval{+3.25} \\
\bottomrule
\end{tabular}%
}
\end{table}

\subsection{Ablation Studies}
\label{sec:ablation}

To understand which design choices contribute to \sysname{}'s performance, we ablate three components on SLIP-1B across all five tasks (Table~\ref{tab:ablation}): \textbf{latent projection}, which maps continuous thoughts back into the input embedding space for the next reasoning step; \textbf{reasoning distillation}, which enables the student to acquire the teacher's reasoning ability without generating explicit CoT; and \textbf{masking}, which regularizes training.

Removing the latent projection $\pi$ causes the largest drop, reducing average accuracy from 68.38\% to 62.23\% ($-6.15\%$). This shows that learning a transformation between the hidden states produced by one reasoning step and the inputs consumed by the next is critical for effective latent reasoning. Removing distillation while retaining the latent reasoning steps reduces average accuracy to 64.34\% ($-4.04\%$). Without distillation, the continuous thoughts receive no supervision and the model is trained with only the final-answer cross-entropy loss, making it functionally equivalent to a No-CoT model with additional forward passes. The drop confirms that supervising the continuous thoughts through distillation is necessary for the model to learn effective latent representations. Finally, removing masking reduces average accuracy to 66.58\% ($-1.80\%$), suggesting it prevents the model from bypassing the latent reasoning steps.

\begin{table}[h]
\centering
\caption{\textbf{\sysname{} ablation experiments} with the SLIP-1B backbone on the five reasoning benchmarks. We report answer accuracy (\%). Default settings are marked in \colorbox{headergreen}{green}.
}
\label{tab:ablation}
\resizebox{0.8\linewidth}{!}{%
\begin{tabular}{l ccccc r l}
\toprule
Method & ECG & Sleep & HAR & TSR & Engine & Avg. & $\Delta$ \\ %
\midrule
\rowcolor{headergreen}
\sysname{} (SLIP-1B) (Ours)            & 82.74 & 78.12 & 74.84 & 67.32 & 38.86 & 68.38 & \\ %
\quad w/o projection $\pi$        & 80.87 & 77.68 & 73.60 & 50.00 & 29.02 & 62.23 & \negval{-6.15} \\ %
\quad w/o reasoning distillation & 78.54 & 77.25 & 74.45 & 56.23 & 35.23 & 64.34 & \negval{-4.04} \\ %
\quad w/o mask              & 79.41 & 78.44 & 72.18 & 69.17 & 33.68 & 66.58 & \negval{-1.80} \\ %
\bottomrule
\end{tabular}%
}
\end{table}

\section{Conclusion}

\textbf{Limitations and future work.}
Our work has several limitations. First, \sysname{} still relies on explicit CoT annotations for training as they are required by the teacher, restricting its use to settings where such data can be obtained or synthetically generated. Second, the number of latent reasoning steps is fixed at $K=6$ regardless of task or input complexity. Learning to adaptively allocate computation per input, for instance through reinforcement learning over the number of continuous thoughts, is a natural extension. Third, on our smallest backbone (ITFormer-0.5B), all reasoning methods, including \sysname{}, underperform No-CoT in average F1, indicating that multi-step reasoning may require a minimum model capacity to be effective. Understanding how this capacity threshold interacts with model architecture and task complexity remains an open question.

\textbf{Conclusion and broader impact.}
Our experiments show that finetuning TSLMs with explicit CoT reasoning does not consistently improve average F1 over finetuning with direct answers alone. We hypothesize that this is due to the verbalization bottleneck.
Specifically, when a TSLM must verbalize a continuous time series before reasoning over it, subsequent reasoning steps are conditioned on incomplete or inaccurate textual descriptions rather than on the original signal, which degrades downstream performance. 
\sysname{} addresses this by performing reasoning in the model's joint time series--language space instead of in text. Across four backbones from 0.5B to 7B parameters, we find that \sysname{} improves average F1 over CoT finetuning by up to $10.79\%$ (ChatTS-7B), while reducing the number of generated tokens by up to $95.8\%$. Qualitative analyses further show that the continuous thoughts produced by \sysname{} can be decoded into readable textual traces through standard language decoding. More broadly, as TSLMs are increasingly applied to safety-critical domains such as clinical monitoring and industrial diagnostics, \sysname{} offers a post-training paradigm that improves both reasoning accuracy and inference efficiency without sacrificing interpretability.

\clearpage
\subsection*{AI use statement}

In this work, we used generative AI tools to aid and polish writing, for retrieval and discovery (e.g., finding related work), and for generating synthetic datasets. We have reviewed all AI-assisted work. Specifically, all AI-polished text was manually reviewed and revised by the authors to ensure accuracy and faithful representation of our contributions. Retrieved references were verified against original sources, and all synthetic datasets were validated for correctness before use in experiments. We did not use generative AI tools for research ideation or execution, to draft sections of the paper, or to prove mathematical claims. We take responsibility for the final content of this work, including text, claims, and artifacts produced with the aid of generative AI.

\subsection*{Acknowledgments}

The authors acknowledge support for this research from \textbf{Evergreen: A Generative AI and Behavioral Sensing Digital Ecosystem to Promote Student Wellness and Flourishing.} This work is made possible through philanthropic gifts to Dartmouth College dedicated to advancing AI-supported well-being and flourishing of college students.

\bibliographystyle{assets/plainnat}
\bibliography{iclr2027_conference}

\clearpage
\beginappendix
\section{Dataset Details}
\label{sec:app_dataset}

We construct a reasoning trace for every task, since \sysname{} is trained by distilling one. The five tasks divide by how that trace is obtained: ECG and Engine carry certified, stage-wise ground-truth labels, so their traces are assembled deterministically from those labels; HAR, Sleep, and TSR do not, so their traces are distilled from a teacher model. In all cases the trace is used only to shape training and is never emitted at inference. Table~\ref{tab:datasets} reports the split sizes, channel counts, series lengths, and answer format of each task.

\paragraph{ECG.}
The original ECG-Reasoning benchmark is multi-stage: each reasoning step is annotated separately and posed as an independent question. We reformat it into a single binary diagnosis question (e.g., presence vs.\ absence of atrial fibrillation) by concatenating the per-stage gold answers into one trace that proceeds from criterion selection, to finding identification, to ECG grounding, and finally to the diagnostic decision, which serves as the answer.

\paragraph{Engine.}
Engine is an aero-engine sensor dataset. We keep only the subset of samples that carry a certified severity label corresponding to each sensor channel, since only these samples support a verifiable ground-truth trace. From these certified labels the rationale is generated deterministically as a multi-step trace: a per-component roll-call over the five rotating components, a severity read-off, and the maintenance implication that follows from the resulting (component, severity) pair.

\paragraph{HAR and Sleep.}
HAR and Sleep are a human activity recognition and a sleep stage classification task, respectively; both require reasoning over time series to answer questions about the underlying activity or sleep stage. Following \citet{artist}, we generate the supervised finetuning rationales through a distillation pipeline. We prompt Qwen3.5-27B to answer each question while conditioning it on rendered plots of the signal, the summary statistics, and the gold answer. The model is first asked to identify at most three contiguous segments that are essential for answering the question, and then to produce a step-by-step explanation that reaches the answer using only those segments and the statistics. For each segment, the rationale for selecting it is stated separately from the analysis of what the segment reveals, which keeps the segment selection decoupled from its interpretation. We transform the resulting responses into structured traces that interleave natural language reasoning with explicit segment-selection calls, then post-process them to check correctness and adherence to the template.

\paragraph{TSR.}
TSR is the counterfactual subset of the original TSR benchmark~\citep{merrill-etal-2024-language}, a multiple-choice reasoning task over mixed-domain real-world signals (e.g., residential solar generation or city temperature). Given a signal, the model must predict a real-world consequence or judgment whose correct answer flips with the signal's qualitative behavior. Each item pairs an original series with its counterfactual counterpart; the two share the same question and answer choices but differ in the time series and in the correct answer, so the answer can only be determined by reading the series. As the dataset does not provide a reasoning rationale, we generate one with a distillation pipeline following \citet{artist}, as for HAR and Sleep.

\begin{table}[h]
\centering
\caption{Overview of datasets.}
\label{tab:datasets}
\small
\begin{tabular}{lrrrrl}
\toprule
Dataset & Train & Test & Channels & Len & Format \\
\midrule
ECG~\citep{ECG-Reasoning-Benchmark}      & 5{,}800  & 643   & $12$ & $1000$         & Binary (Yes/No) \\
HAR~\citep{opentslm}             & 68{,}542 & 8{,}222 & $3$ & $128$          & Multiple choice \\
Sleep~\citep{opentslm}           & 7{,}413  & 923   & $1$ & $1500$           & Multiple choice \\
TSR~\citep{merrill-etal-2024-language} & 22{,}566 & 4{,}094 & $1$ & $128\text{--}1020$ & Multiple choice \\
Engine~\citep{itformer}          & 2{,}305  & 193   & $33$ & $600$           & Multiple choice \\
\bottomrule
\end{tabular}
\end{table}

\section{Backbone Architectures}
\label{sec:app_backbones}
\begin{table}[h]
\centering
\caption{\textbf{The four backbones along three design axes.} $|H_s|$ is the number of sensor
tokens handed to the language model; $C$ is the channel count, $L$ the series length, and $P$ the
patch size of Table~\ref{tab:hparams}. \emph{Concat} places $H_s$ in the token sequence and reads
it with self-attention; \emph{cross-attn} keeps it outside and injects it through dedicated layers.}
\label{tab:backbones}
\small
\setlength{\tabcolsep}{4pt}
\begin{tabular}{llllll}
\toprule
Backbone & LM & Sensor encoder & Connector & $|H_s|$ & Fusion (depth) \\
\midrule
ChatTS-7B          & 7B   & per-patch MLP        & none (native space)          & $C\lceil L/P\rceil$ & concat (all layers) \\
ITFormer-0.5B      & 0.5B & PatchTST             & question-cond.\ resampler    & $25$ & concat (all layers) \\
SLIP-1B            & 1B   & language-pretrained  & attention pooler             & $64$ & cross-attn (last $4$) \\
OpenTSLM-1B        & 1B   & PatchTST             & Perceiver resampler          & $64$ & cross-attn (every layer) \\
\bottomrule
\end{tabular}
\end{table}
All four backbones follow a common three-stage template. A pretrained language
model $f_\phi$ is extended to a second modality by an encoder $g_\psi$ that maps the series $X_s$ to
features $Z_s$, a connector $h_\omega$ that maps these features into the language embedding space as
$H_s = h_\omega(Z_s)$, and a fusion mechanism that determines where inside $f_\phi$ the sensor tokens
are read. All four patch the input. A length-$L$, $C$-channel recording is divided into windows of
$P$ timesteps, analogous to image patches. The backbones differ along three axes: the patch encoding,
the sensor token count, and the fusion mechanism. Table~\ref{tab:backbones} summarizes the four
configurations.

\paragraph{Encoder.}
ChatTS~\citep{chatts} min-max normalizes each series, writes the scale and offset into the prompt,
and applies a five-layer MLP to each patch in isolation, so cross-time interaction occurs only inside
the language model. ITFormer~\citep{itformer} uses a PatchTST encoder, pretrained separately and
frozen, that attends across the patches of a channel, so its features carry local temporal context.
OpenTSLM~\citep{opentslm} embeds each patch with a jointly trained 1D convolution, one channel at a
time, and defers temporal mixing to the connector. SLIP~\citep{slip} uses an encoder pretrained on
large-scale sensor caption pairs with contrastive and captioning objectives, with a FlexMLP that
adapts patch granularity to the sampling rate, so its features are aligned to language before QA
finetuning.

\paragraph{Connector.}
ChatTS uses no connector. The MLP emits vectors in the language embedding space, and the patches of
each series are inserted at the position where the series is named in the prompt. Every patch of
every channel becomes one token, so $|H_s| = C \lceil L/P \rceil$ grows with the channel count and
the series length. Table~\ref{tab:hparams} therefore uses a larger patch size on the $12$-channel
ECG and $33$-channel Engine tasks, where $P=8$ yields over a thousand sensor tokens and
exceeds the memory available at the 7B scale. The other three backbones use a resampler, a fixed set
of learned queries that cross-attend to the encoder features and return one token each. The token
count is fixed by the architecture and independent of the series length, $25$ for ITFormer and $64$
for SLIP and OpenTSLM. ITFormer prepends $25$ learnable instruct tokens to the question, refines them
with self-attention, and uses them as queries that pool the encoder features across channels and then
across timesteps, so the returned tokens depend on the instruction. SLIP and OpenTSLM use
question-agnostic resamplers, a Perceiver resampler in OpenTSLM and caption queries pretrained with
the encoder in SLIP. The $64$ tokens are computed from the series alone, and the question interacts
with them only when the text tokens cross-attend to $H_s$ inside $f_\phi$.

\paragraph{Fusion.}
ChatTS and ITFormer concatenate the sensor tokens into the prompt, $H = f_\phi([H_q; H_s])$. The
signal is read by self-attention at every layer, no parameters are added to the language model, and
the two modalities share one attention budget. SLIP and OpenTSLM keep $H_s$ outside the sequence and
inject it through gated cross-attention layers interleaved into $f_\phi$, $H = f_\phi(H_q \mid H_s)$,
following Flamingo~\citep{alayrac2022flamingovisuallanguagemodel}. The sensor pathway has separate
parameters and a separate attention budget, a tanh gate initialized at zero preserves the pretrained
text behavior at initialization, and memory is constant in the series length. OpenTSLM injects at
every layer. SLIP injects at the final four layers, so lower layers process the question as text
alone.

\paragraph{Coverage.}
The four backbones span all three axes and language models from $0.5$B to $7$B parameters.
\sysname{} modifies none of these components and changes only the tokens the language model emits
between the prompt and the answer, so evaluating on all four separates its effect from any particular
integration of the signal. The attention analysis of
Appendix~\ref{sec:app_attn} requires a concatenation backbone, because $\lambda_{\mathrm{ts}}$ is defined as a share
of a single self-attention row and is measurable only when the sensor tokens are in the sequence; we use
ChatTS-7B as a representative one. The
latent loop of Equation~\ref{eq:latent_loop}, by contrast, applies to all four backbones. Each continuous thought
$z_k$ re-enters $f_\phi$ as an input embedding and reads the signal through the fusion pathway of the
backbone, whereas an explicit chain conditions increasingly on its own generated text.

\section{Architecture Variant Discussion}
\label{sec:app_arch}
We briefly explore whether the effect of the finetuning method changes across architecture families. OpenTSLM has released two controlled variants, OpenTSLM-Flamingo and OpenTSLM-SoftPrompt, which share the same $1$B language model but differ in how the sensor tokens interact with it (Appendix~\ref{sec:app_backbones}). This pair gives us an opportunity for a controlled experiment that isolates the fusion pathway while keeping all other factors fixed.

As shown in Table~\ref{tab:arch}, OpenTSLM-SoftPrompt outperforms OpenTSLM-Flamingo in average accuracy under every finetuning method, though which variant leads on any given task varies without a consistent pattern. \sysname{} improves accuracy on both variants and narrows the gap between them from $3.57$\% under No-CoT to $0.67$\%, suggesting that latent reasoning reduces the model's sensitivity to how sensor tokens are fused with the language backbone. Explicit chain-of-thought, by contrast, does not produce a clear benefit on either variant.

Overall, the gains from latent reasoning are consistent across both fusion pathways, and the choice between them matters less under \sysname{} than under No-CoT. However, whether this behavior holds for larger language backbones (e.g., 7B) remains open for future work.
\begin{table}[h]
\centering
\caption{\textbf{Effect of the fusion pathway.} Accuracy and Macro-F1 (\%) on the two OpenTSLM
variants, which share a $1$B language model and differ only in fusion. Ours in
\colorbox{headergreen}{green}; \textbf{bold} and \underline{underline} are best and second-best per
column within a variant; $\Delta$ is the change in average over No-CoT.}
\label{tab:arch}
\resizebox{\linewidth}{!}{%
\begin{tabular}{ll cc cc cc cc cc rl rl}
\toprule
 & & \multicolumn{2}{c}{ECG} & \multicolumn{2}{c}{Sleep} & \multicolumn{2}{c}{HAR} & \multicolumn{2}{c}{TSR} & \multicolumn{2}{c}{Engine} & \multicolumn{4}{c}{Avg.} \\
\cmidrule(lr){3-4} \cmidrule(lr){5-6} \cmidrule(lr){7-8} \cmidrule(lr){9-10} \cmidrule(lr){11-12} \cmidrule(lr){13-16}
Model & Finetune Method & Acc & F1 & Acc & F1 & Acc & F1 & Acc & F1 & Acc & F1 & Acc & $\Delta$ & F1 & $\Delta$ \\
\midrule
\multirow{5}{*}{OpenTSLM-SoftPrompt}
 & No-CoT & 61.59 & 60.98 & \underline{84.29} & \underline{73.93} & 73.46 & 69.85 & 49.15 & 48.65 & 31.09 & 30.64 & 59.92 & & 56.81 & \\
 & CoT & 52.10 & 51.94 & 83.42 & 72.87 & 71.14 & 67.00 & 53.03 & 53.02 & 32.12 & 32.11 & 58.36 & \negval{-1.55} & 55.39 & \negval{-1.42} \\
 & iCoT & \underline{63.76} & 63.62 & 84.07 & 72.10 & 72.55 & 69.34 & 54.86 & 54.85 & 26.42 & 18.03 & 60.33 & \posval{+0.41} & 55.59 & \negval{-1.22} \\
 & COCONUT & \textbf{65.16} & \underline{64.48} & 83.42 & 71.25 & \underline{74.18} & \textbf{71.18} & \underline{56.74} & \underline{56.50} & \underline{34.72} & \underline{33.83} & \underline{62.84} & \posval{+2.93} & \underline{59.45} & \posval{+2.64} \\
 & \hl \sysname{} (Ours) & \hl\textbf{65.16} & \hl\textbf{64.80} & \hl\textbf{85.81} & \hl\textbf{75.62} & \hl\textbf{74.25} & \hl\underline{70.94} & \hl\textbf{58.04} & \hl\textbf{58.00} & \hl\textbf{37.31} & \hl\textbf{36.08} & \hl\textbf{64.11} & \hl\posval{+4.20} & \hl\textbf{61.09} & \hl\posval{+4.28} \\
\midrule
\multirow{5}{*}{OpenTSLM-Flamingo}
 & No-CoT & 71.07 & 70.90 & \underline{77.57} & \underline{65.25} & \underline{71.31} & \underline{66.22} & 33.29 & 33.32 & \underline{28.50} & 11.09 & 56.35 & & 49.36 & \\
 & CoT & 65.47 & 65.33 & 72.81 & 58.02 & 67.61 & 61.71 & \underline{52.69} & \underline{53.13} & 27.98 & \underline{26.48} & 57.31 & \posval{+0.96} & 52.93 & \posval{+3.58} \\
 & iCoT & \underline{72.94} & \underline{72.78} & \underline{77.57} & 65.22 & 69.98 & 65.40 & 51.10 & 51.08 & 26.94 & 24.20 & \underline{59.71} & \posval{+3.36} & \underline{55.74} & \posval{+6.38} \\
 & COCONUT & 70.45 & 70.38 & 76.27 & 62.80 & 70.82 & 65.89 & 47.92 & 41.91 & 24.87 & 16.84 & 58.07 & \posval{+1.72} & 51.56 & \posval{+2.21} \\
 & \hl \sysname{} (Ours) & \hl\textbf{73.56} & \hl\textbf{73.49} & \hl\textbf{80.72} & \hl\textbf{68.91} & \hl\textbf{72.40} & \hl\textbf{68.20} & \hl\textbf{58.40} & \hl\textbf{58.44} & \hl\textbf{32.12} & \hl\textbf{27.51} & \hl\textbf{63.44} & \hl\posval{+7.09} & \hl\textbf{59.31} & \hl\posval{+9.95} \\
\bottomrule
\end{tabular}%
}
\end{table}

\section{Implementation Details}
\label{implementation}
\begin{table}[h]
\centering
\caption{\textbf{Per-task hyperparameters} for the four backbones and the OpenTSLM-SoftPrompt variant. One row covers all five
finetuning methods, which share every setting except the distillation weight $\lambda$ of
\sysname{}.}
\label{tab:hparams}
\small
\setlength{\tabcolsep}{5pt}
\begin{tabular}{ll ccccc}
\toprule
Backbone & Dataset & Patch Size & Epochs & lr & wd & $\lambda$ \\
\midrule
\multirow{5}{*}{ChatTS-7B}
 & ECG      & 32    & 20 & 1e-5 & 1e-2 & 10 \\
 & Sleep           & 8     & 20 & 1e-5 & 1e-2 & 10 \\
 & HAR             & 8     & 4  & 1e-5 & 1e-2 & 10 \\
 & TSR & 8     & 4  & 1e-5 & 1e-2 & 1  \\
 & Engine          & 50    & 5  & 1e-5 & 1e-2 & 20 \\
\midrule
\multirow{5}{*}{SLIP-1B}
 & ECG      & 32    & 20 & 5e-5 & 1e-2 & 10 \\
 & Sleep           & 64    & 20 & 1e-4 & 1e-2 & 10 \\
 & HAR             & 4     & 4  & 1e-4 & 1e-2 & 10 \\
 & TSR & 16/32 & 4  & 1e-4 & 1e-2 & 10 \\
 & Engine          & 32    & 5  & 1e-5 & 1e-2 & 20 \\
\midrule
\multirow{5}{*}{OpenTSLM-Flamingo}
 & ECG      & 4     & 20 & 1e-4 & 1e-2 & 10 \\
 & Sleep           & 4     & 20 & 1e-4 & 1e-2 & 1  \\
 & HAR             & 4     & 4  & 1e-4 & 1e-2 & 1  \\
 & TSR & 4     & 4  & 1e-5 & 1e-2 & 10 \\
 & Engine          & 4     & 5  & 1e-5 & 1e-2 & 10 \\
\midrule
\multirow{5}{*}{OpenTSLM-SoftPrompt}
 & ECG      & 100   & 20 & 1e-4 & 1e-2 & 10 \\
 & Sleep           & 4     & 20 & 1e-4 & 1e-2 & 1 \\
 & HAR             & 4     & 4  & 1e-4 & 1e-2 & 1 \\
 & TSR & 4     & 4  & 1e-5 & 1e-2 & 10  \\
 & Engine          & 50    & 5  & 1e-5 & 1e-2 & 10  \\
\midrule
\multirow{5}{*}{ITFormer-0.5B}
 & ECG      & 60    & 20 & 1e-5 & 1e-2 & 10 \\
 & Sleep           & 60    & 20 & 1e-5 & 1e-2 & 10 \\
 & HAR             & 60    & 4 & 1e-5 & 1e-2 & 10 \\
 & TSR & 60    & 4  & 1e-5 & 1e-2 & 10 \\
 & Engine          & 60    & 5  & 1e-5 & 1e-2 & 10 \\
\bottomrule
\end{tabular}
\end{table}
\paragraph{Training configuration.}
All runs use LLaMA-Factory\footnote{\url{https://github.com/hiyouga/LLaMA-Factory}} with the AdamW optimizer, a cosine learning-rate schedule with warmup ratio 0.02, bf16 precision, DeepSpeed ZeRO-2, gradient clipping at 1.0, and an effective batch size of 128. In all settings, the TSLM is fully finetuned. The \sysname{} runs use the default configuration with the latent projector enabled, $K=6$ continuous thoughts, and a mask ratio of 0.3 on patchified sensor streams during training (Section~\ref{sec:method}). Table~\ref{tab:hparams} reports the settings that vary across backbones and tasks.

We start from the recommended hyperparameters of each backbone~\citep{chatts,itformer,opentslm,slip} and make only the adjustments needed for the training loss to flatten. Within each backbone, the learning rate and weight decay are held fixed across all five finetuning methods. The only method-specific setting is the distillation weight $\lambda$ of \sysname{}. We default to $\lambda=10$ following CODI~\citep{codi} and adjust it within $\{1, 10, 20\}$ so that the weighted distillation loss stays on the same order of magnitude as the student and teacher cross-entropy losses.

Patch sizes follow each backbone's default. OpenTSLM-Flamingo, OpenTSLM-SoftPrompt, ChatTS, and ITFormer use fixed patch sizes of 4, 4, 8, and 60, respectively. We slightly adjust them for computational efficiency. For ChatTS, we increase the patch size on ECG and Engine to 32 and 50 to avoid memory overflow at the 7B scale. For OpenTSLM-SoftPrompt, we increase the patch size on the same two datasets to 100 and 50 so that its effective batch size matches that of OpenTSLM-Flamingo. SLIP uses an adaptive patch size per dataset, following its own recommendation.

For each dataset, the epoch count is shared across all backbones and all five finetuning methods, and we report the final checkpoint with no checkpoint selection, so no test-set information enters the schedule. Evaluation uses greedy decoding with at most 512 new tokens for every method. Each run uses 4$\times$H200.

\clearpage
\section{Per-Token Predictive Entropy via the Logit Lens}
\label{sec:logit-lens-entropy}
\textbf{Setup.} To characterize \emph{what a continuous thought encodes}, we decode every
intermediate state through the model's own output head, following the logit-lens analysis of
latent reasoning \citep{deng2026llmlatentreasoningchain}. A language model maps any hidden
state $h_t$ to a next-token distribution with the same unembedding it uses at the output
layer, $p_t = \mathrm{softmax}(W_U h_t/\tau)$, where $W_U \in \mathbb{R}^{V\times d}$ is the
(tied) language-model head, $V$ the vocabulary size, and $\tau>0$ the temperature. We quantify
how concentrated this distribution is with the Shannon entropy in bits, $H(p_t) =
-\sum_{v=1}^{V} p_t(v) \log_2 p_t(v)$. A low $H(p_t)$ indicates that the state places nearly
all probability mass on a single token, whereas a high $H(p_t)$ indicates that the state
distributes mass across multiple candidate continuations and has not yet committed to one of
them.

The readout is identical in both decoding modes, and only the input to $W_U$ differs. In
explicit decoding, $h_t$ is computed after a discrete token has been embedded and fed back, so
the $\arg\max$ of $p_t$ is the emitted token and the lens recovers a prediction the model has
already committed to. In latent decoding, the model emits no token during reasoning. Each
continuous thought $z_k$ is the last-layer hidden state, which is mapped through
$\pi(\cdot)$ and fed back as the input embedding for the next step. The distribution
$p_{z_k}=\mathrm{softmax}(W_U z_k/\tau)$ is therefore a \emph{diagnostic} readout that
indicates which vocabulary tokens the thought would favor if it were decoded at that step.
Because the thought is never discretized, a single latent step can retain probability mass on
several candidate tokens rather than on one. We summarize this spread by the \emph{effective
number of candidate tokens} per thought, $N_{\text{eff}} = 2^{\bar{H}}$, where $\bar{H}$ is the
mean of $H(p_{z_k})$ over the $K$ thoughts. $N_{\text{eff}}$ equals the number of equiprobable
tokens whose entropy matches $\bar{H}$, and $N_{\text{eff}}=1$ corresponds to a deterministic
prediction. For each example, we record $H(p_t)$ and the sorted top-$k$ probabilities of $p_t$
at every position, and we define the \emph{commitment point} as the first answer token whose
top-1 probability is at least $0.95$.
\begin{algorithm}[H]
\caption{Logit-lens entropy readout for explicit and latent decoding.}
\label{alg:logit-lens-entropy}
\lstset{style=mocov3}
\begin{lstlisting}[language=python, escapechar=@, label=code:logit-lens-entropy]
import torch, torch.nn.functional as F

# W_U: LM head, tau: temperature, h: last-layer state at last position
def logit_lens(h, tau=1.0):
    return F.softmax(W_U(h).float() / tau, dim=-1)   # (V,)

def entropy_bits(p):
    return -(p * p.clamp_min(1e-12).log2()).sum()

def record(h, tag):                        # log entropy, top-k, phase
    p = logit_lens(h)
    H_seq.append(entropy_bits(p)); heat.append(p.topk(30).values); phase.append(tag)
    return p

def decode_answer(h):                      # greedy token decoding
    for _ in range(max_new_tokens):
        tok = record(h, "answer").argmax()
        if tok == eos: break
        h = step(embed(tok))[-1]

# ---- CoT ----
H_seq, heat, phase = [], [], []
decode_answer(encode(prompt)[-1])

# ---- latent reasoning ----
H_seq, heat, phase = [], [], []
h = encode(prompt + [bot])[-1]
for k in range(K):
    record(h, "latent")                    # h is thought z_k
    h = step(pi(h))[-1]
h = step(embed(eot))[-1]                   # end thinking
decode_answer(h)

# effective number of candidate tokens per continuous thought
H_lat = torch.stack([H for H, ph in zip(H_seq, phase) if ph == "latent"])
N_eff = 2 ** H_lat.mean()

# commitment point: first answer token with top-1 prob >= 0.95
commit = next((i for i, ph in enumerate(phase)
               if ph == "answer" and heat[i][0] >= 0.95), None)
\end{lstlisting}
\end{algorithm}

\begin{figure}[h]
    \centering
    \includegraphics[width=0.9\linewidth]{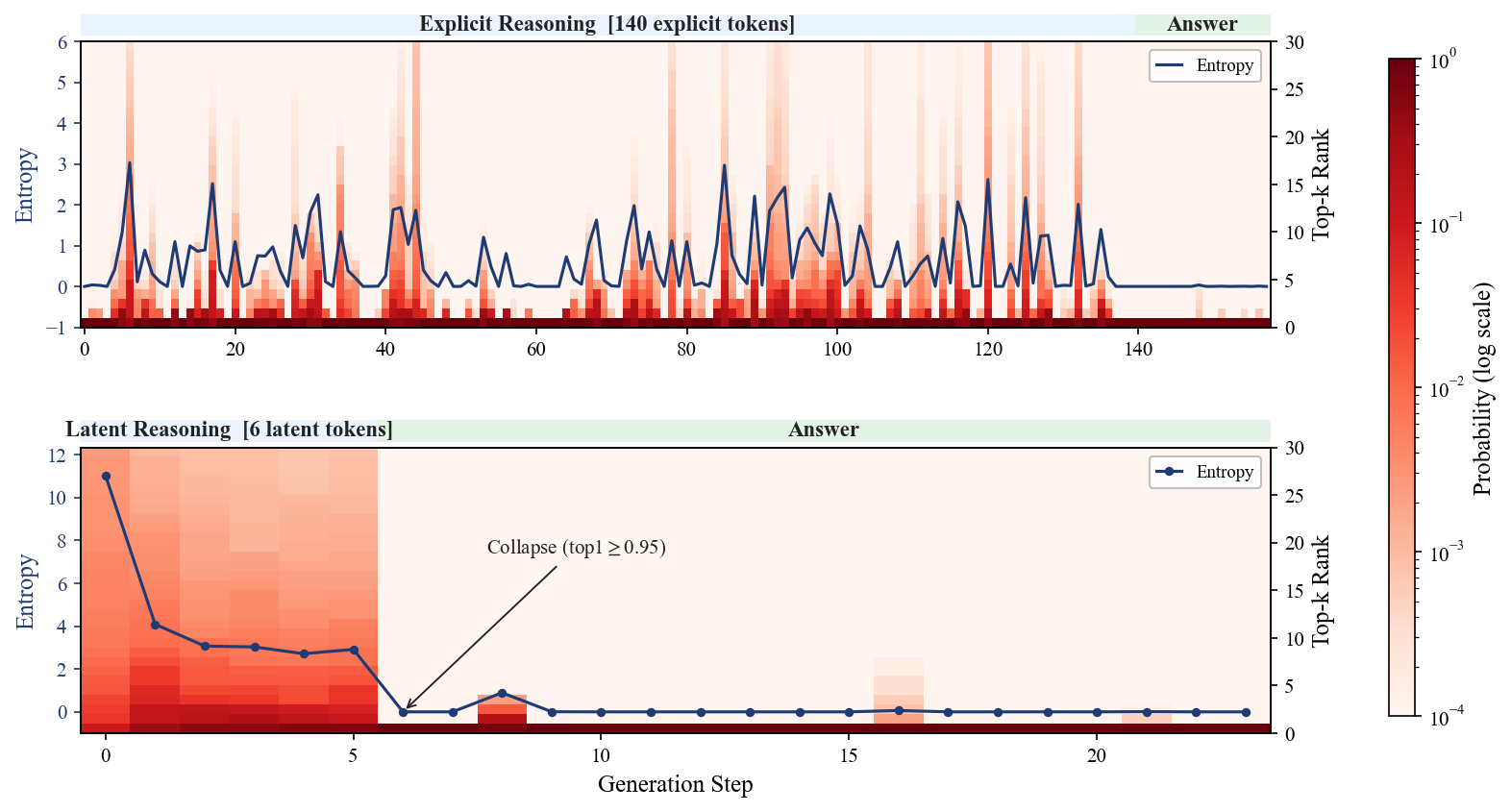}
    \caption{\textbf{Per-step logit-lens readout on one TSR example.} At each
    step, we plot the next-token distribution read through the logit lens, with tokens ranked
    by probability and colored on a log scale, together with its entropy (blue).
    \textbf{Top:} under CoT the distribution is sharply peaked at nearly all of the $140$
    steps. \textbf{Bottom:} under latent reasoning the $6$ continuous thoughts spread probability
    over many tokens ($N_{\text{eff}}\approx 22$), entropy decreases after the first thought,
    and the distribution reaches the commitment point ($\text{top-}1\ge 0.95$) only at the
    first answer token.}
    \label{fig:logit-lens-heatmap}
\end{figure}

\clearpage
\paragraph{Result.} Figure~\ref{fig:logit-lens-heatmap} compares CoT and latent reasoning on
the same example. Under CoT (top), the readout has low entropy at nearly all of the $140$
steps. Each emitted token is already committed when it is produced, so the model reaches the
answer through a long sequence of near-deterministic steps. Under latent reasoning (bottom),
the computation is concentrated in $6$ continuous thoughts whose readouts spread probability over
many vocabulary tokens ($N_{\text{eff}}\approx 22$), which indicates that each thought retains
mass on several candidate continuations rather than committing to a single token. Entropy
decreases after the first thought, and the distribution reaches the commitment point
($\text{top-}1\ge 0.95$) only at the first answer token. The continuous thoughts therefore do not
reproduce a compressed version of the CoT trajectory, and instead follow an entropy profile
that differs from explicit decoding.

\clearpage

\section{Attention on the Time Series}
\label{sec:app_attn}
A model can score well on multiple-choice temporal QA by exploiting linguistic priors over the
answer choices instead of reading the signal, and accuracy alone says little about which part of
the prompt each format actually works over.
To shed light on this, we investigate the attention weight distributions during generation.
\paragraph{Setting.} We run the attention analysis in this section on ChatTS-7B, a concatenation backbone, since SLIP and OpenTSLM inject the series through
cross-attention, where the allocation is not measurable. We take a representative TSR
example and average $\lambda_{\mathrm{ts}}$ over the reasoning-token queries of each format, namely
the answer token for No-CoT, the chain-of-thought span for CoT, and
the latents for latent reasoning.

\paragraph{Attention Maps.}
We visualize the model's layer-wise log-scaled attention vectors following the protocol of \citet{xiao2024efficientstreaminglanguagemodels}. A row is a generated token and a column is a
token it can look back at, so the maps are lower triangular. The \textcolor{cTimeSeries}{time series} column is brightest under latent reasoning at the
input and fades in the later layers, so the signal is absorbed early and the deepest layers work
over text. The \textcolor{cReasoning}{reasoning} column is wide in panel~(b) and almost absent in
panel~(c), so the mass that leaves the time series column under CoT is the mass its
own trace takes. Latent reasoning has no such trace to fall back on and therefore keeps
the series in view through the depths where the computation happens.

\paragraph{Attention Allocation.} To quantify the allocation, we follow \citet{chen2024imageworth12tokens} and measure \textit{the total attention score one type of token receives in one layer}. Let $\alpha^{(\ell)}_{qj}$ be the post-softmax attention
that a generated token $q$ pays to a key $j$ at layer $\ell$, averaged over heads, and let $S$ be
the set of time series tokens. The allocation of the series is
$$
\lambda^{(\ell)}_{\mathrm{ts}} \;=\; \sum_{j \in S,\; j < q} \alpha^{(\ell)}_{qj},
\qquad \text{where} \quad \sum_{j \le q} \alpha^{(\ell)}_{qj} = 1 .
$$
Since each row sums to one, $\lambda_{\mathrm{ts}}$ is the fraction of a token's attention budget
that goes to the signal, and the rest goes to the other spans it can look back at.

\begin{table}[t]
\centering
\caption{\textbf{Attention on the time series} with ChatTS-7B. Allocation $\lambda_{\mathrm{ts}}$ (\%) is the
share of a reasoning token's attention that lands on the time series tokens, at the first, middle,
and last layers.}
\label{tab:attn_ts}
\small
\begin{tabular}{l rrr}
\toprule
Layer & No-CoT & CoT & \sysname{} \\
\midrule
0 (first)   & 2.95  & 3.39 & 17.35 \\
21 (middle) & 11.18 & 4.65 & 6.54  \\
35 (last)   & 0.72  & 0.27 & 1.21  \\
\bottomrule
\end{tabular}
\end{table}

\begin{figure}[h]
    \centering
    \includegraphics[width=0.9\linewidth]{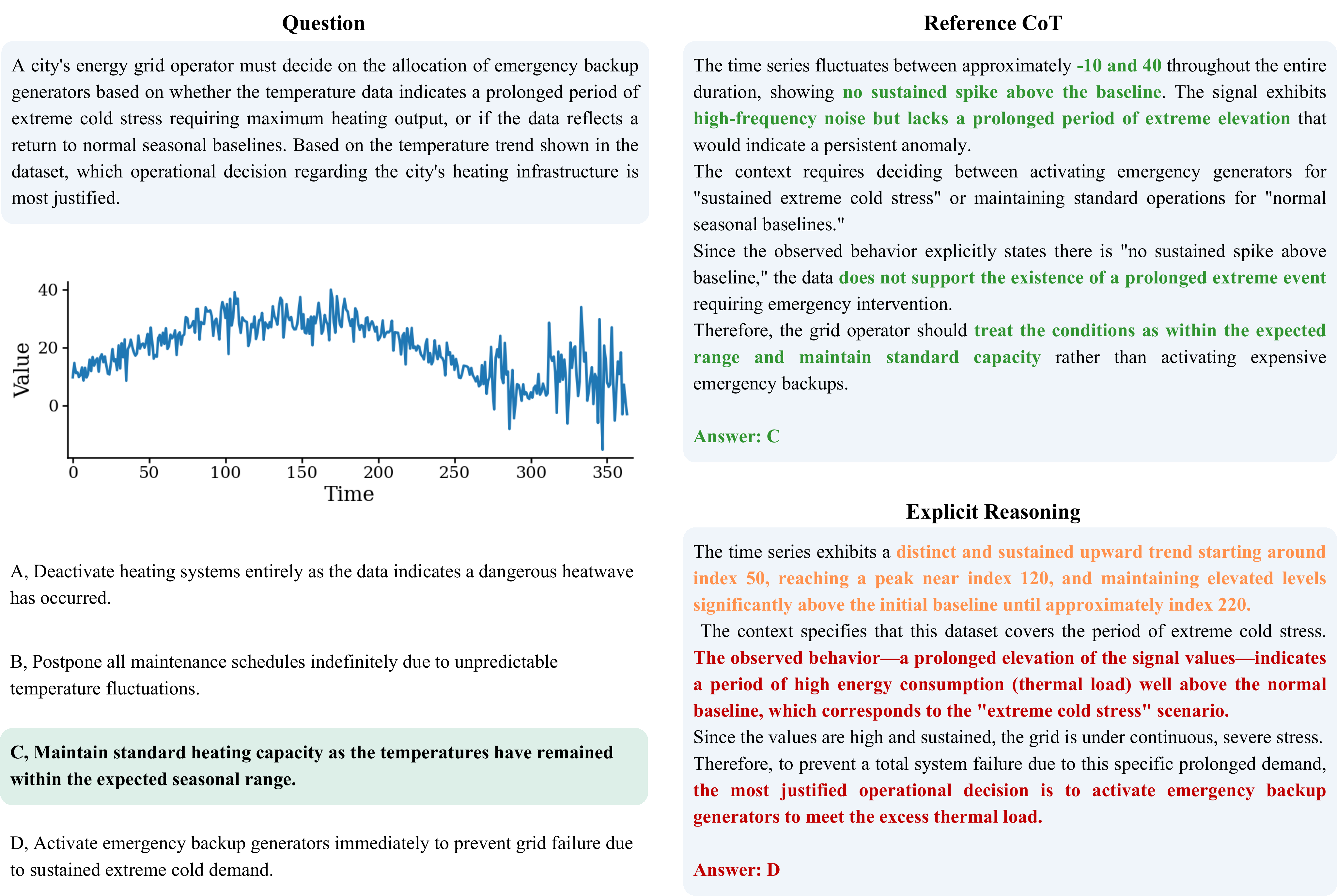}
    \caption{\textbf{Verbalization bottleneck in CoT reasoning.} An operational decision over a temperature series with four choices. \emph{CoT} (bottom right) runs a CoT trace that first verbalizes the signal as text, then reasons from that textual summary rather than the original representation, and reaches an incorrect final answer. Spans are colored \textcolor{green!60!black}{grounded}, \textcolor{orange!90!black}{correct but insufficient}, and \textcolor{red!80!black}{incorrect}. 
    }
    \label{fig:motivation}
\end{figure}

As Table~\ref{tab:attn_ts} shows, latent reasoning allocates the most attention to the time series in the early layers, where it
has direct access to the series, and we read this as the model working out which part of the
series matters given the context and the question. Once the series has been read, the later layers
can move on to the high-level reasoning step, so the allocation drops. No-CoT behaves
differently and peaks in the middle layers. We hypothesize that the model leans on the text
prior in the early layers and only looks back at the series once it has to commit to an answer.
CoT is flat across depth and stays below the other two formats almost everywhere. We
attribute this to the verbalization bottleneck (Figure~\ref{fig:motivation}), which \sysname{} is designed to avoid.

\clearpage
\section{Additional Visualization}
\label{sec:app_vis}
\begin{figure}[h]
    \centering
    \includegraphics[width=\linewidth]{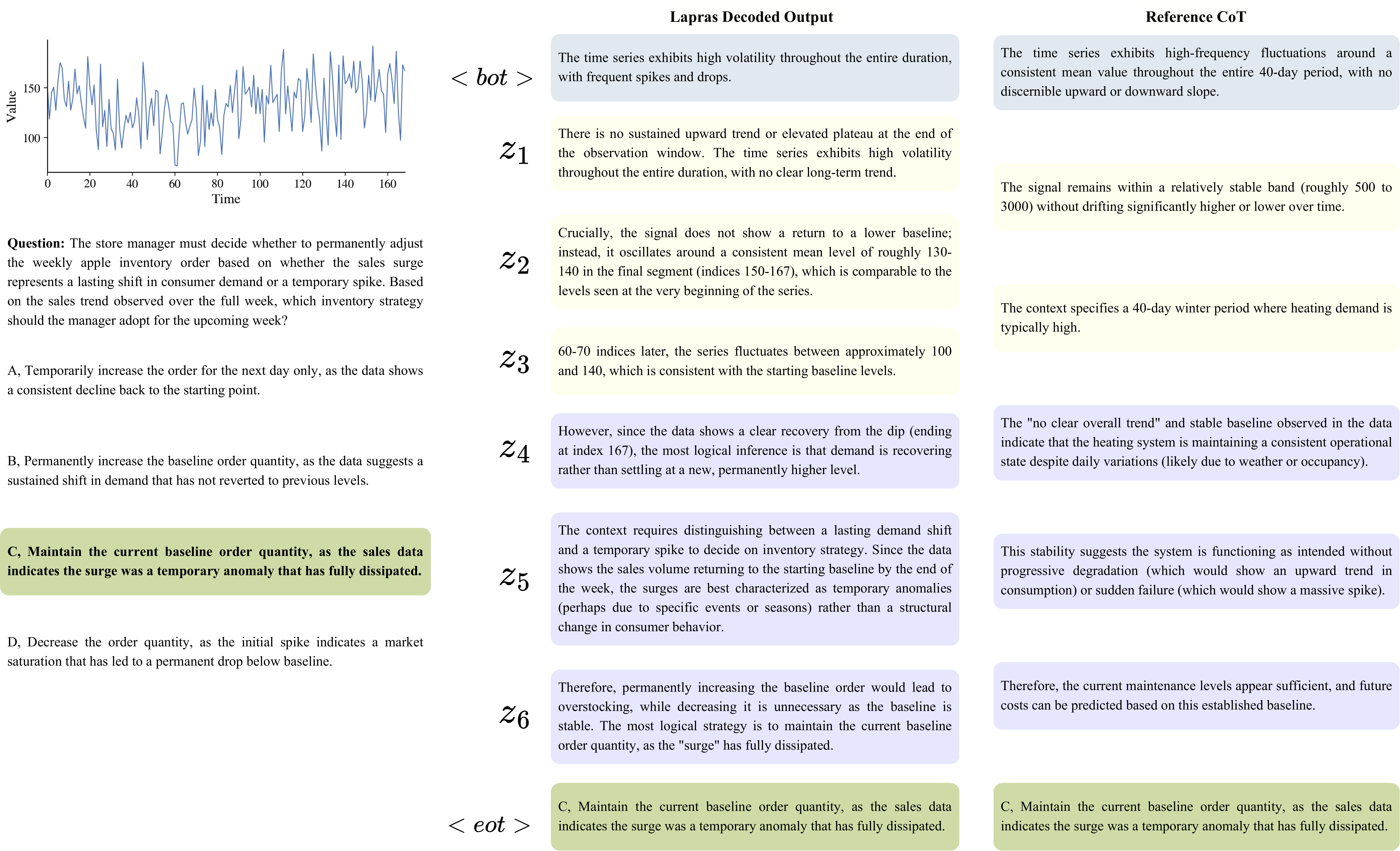}
    \caption{\textbf{Case study of \sysname{}'s interpretability}, obtained by decoding each of the six continuous thoughts $z_1, \dots, z_6$ into text and placing it beside the explicit chain. The decoded steps progress from a \hlfig{cGlobalDesc}{global description} of the series, through \hlfig{cLocalDesc}{local descriptions} of individual segments, to \hlfig{cInference}{inferences} that combine them, and finally to the \hlfig{cAnswerBox}{answer}. Highlight colors match the box fills in the figure.}
    \label{fig:interp_3048}
\end{figure}

\end{document}